\documentclass[sigconf]{acmart}
\usepackage{soul}
\usepackage{url}
\usepackage[utf8]{inputenc}
\usepackage{subcaption}
\usepackage{graphicx}
\usepackage{amsmath}
\usepackage{amsthm}
\usepackage{enumitem}
\usepackage{booktabs}
\usepackage{algorithm}
\usepackage{algorithmic}
\usepackage{textgreek}
\usepackage{multirow}
\usepackage{marvosym}
\usepackage{placeins}
\usepackage{afterpage}
\hypersetup{hidelinks}
\AtBeginDocument{%
  }
\copyrightyear{2026}
\acmYear{2026}
\setcopyright{cc}
\setcctype{by}
\acmConference[MM '26] {Proceedings of the 34th ACM International Conference on Multimedia}{November 10--14, 2026}{Rio de Janeiro, Brazil.}
\acmBooktitle{Proceedings of the 34th ACM International Conference on Multimedia (MM '26), November 10--14, 2026, Rio de Janeiro, Brazil}
\acmISBN{979-8-4007-2213-4/2026/11}
\acmDOI{10.1145/3767308.3835121}

\begin{document}

\title{PSL-RGHA: Parallel Split Learning with Reference Gradient  Half-Space Alignment for Mitigating Gradient Divergence}

\author{Han Li}
\authornote{Both authors contributed equally to this research. \textsuperscript{\Letter}Corresponding authors.}
\email{18850543060@163.com}
\affiliation{%
  \institution{Xiamen University of Technology}
  \city{Xiamen}
  \state{Fujian}
  \country{China}
}

\author{Zheng Lin}
\authornotemark[1]
\email{zhenglin@ieee.org}
\affiliation{%
  \institution{SnT, University of Luxembourg}
  \city{}
  \country{Luxembourg}
}

\author{Miao Yang}
\email{yangmiao@xmut.edu.cn}
\correspondingauthor
\affiliation{%
  \institution{Xiamen University of Technology}
  \city{Xiamen}
  \state{Fujian}
  \country{China}
}

\author{Zehang Lin}
\email{18850975878@163.com}
\affiliation{%
  \institution{Xiamen University of Technology}
  \city{Xiamen}
  \state{Fujian}
  \country{China}
}

\author{Jing Yang}
\email{s2147529@siswa.um.edu.my}
\affiliation{%
  \institution{Universiti Malaya}
  \city{Kuala Lumpur}
  \country{Malaysia}
}

\author{Zihan Fang}
\email{zihanfang3-c@my.cityu.edu.hk}
\affiliation{%
  \institution{City University of Hong Kong}
  \city{Hong Kong}
  \country{China}
}

\author{Qiyang Zhang}
\email{qiyangzhang@pku.edu.cn}
\affiliation{%
  \institution{Peking University}
  \city{Beijing}
  \country{China}
}

\author{Zhaolu Kang}
\email{kangzl9966@gmail.com}
\affiliation{%
  \institution{Peking University}
  \city{Beijing}
  \country{China}
}

\author{Jianmin Li}
\email{lijm@xmut.edu.cn}
\affiliation{%
  \institution{Xiamen University of Technology}
  \city{Xiamen}
  \state{Fujian}
  \country{China}
}

\author{Lin Li}
\email{linli@xmut.edu.cn}
\affiliation{%
  \institution{Xiamen University of Technology}
  \city{Xiamen}
  \state{Fujian}
  \country{China}
}

\author{Mingwei Hong}
\email{mwhong12345@gmail.com}
\affiliation{%
  \institution{Xiamen University of Technology}
  \city{Xiamen}
  \state{Fujian}
  \country{China}
}

\author{Shunzhi Zhu}
\correspondingauthor
\email{szzhu@xmut.edu.cn}
\affiliation{%
  \institution{Xiamen University of Technology}
  \city{Xiamen}
  \state{Fujian}
  \country{China}
}

\renewcommand{\shortauthors}{Han Li et al.}

\begin{abstract}
The escalating complexity of deep neural networks hinders the deployment of federated learning (FL) on resource-constrained devices, especially for multimedia applications such as on-device image classification and speech recognition. 
Parallel split learning (PSL) has emerged as a promising solution by offloading a substantial portion of the computational workload to the server through layer-wise model partitioning, making it particularly suitable for resource-limited visual and speech intelligence scenarios.
However, the absence of client-side aggregation renders PSL susceptible to severe training divergence and potential convergence failure under heterogeneous data distributions.
To address this challenge, we propose PSL-RGHA, a gradient-aligned \underline{PSL} framework that employs \underline{r}eference \underline{g}radient \underline{h}alf-space \underline{a}lignment mechanism to mitigate training divergence among client devices.
PSL-RGHA comprises two novel components: reference gradient progressive construction (RGPC) and gradient half-space alignment (GHA). 
Specifically, RGPC evaluates the directional consistency of client gradients and progressively synthesizes a robust reference gradient via Weiszfeld updates, effectively approximating the global descent direction. 
Then, GHA constrains client gradients within the reference half-space defined by RGPC to mitigate gradient conflicts and rectify local updates. 
Extensive experiments demonstrate that our PSL-RGHA framework outperforms existing baselines in various settings.

\end{abstract}



\maketitle

\section{Introduction}
With the proliferation of the Internet of Things (IoT), billions of interconnected smart devices continuously generate massive privacy-sensitive data~\cite{0-1,0-2}. This unprecedented volume of data has significantly advanced the maturity of machine learning (ML), enabling intelligent IoT applications~\cite{1,peng2024sums,fang2025dynamic,1-1,yuan2024satsense}.
In particular, the rapid growth of multimedia intelligence, which encompasses high-fidelity image recognition and real-time speech processing, has imposed extreme demands on the computational and memory capabilities of edge devices~\cite{26,27,28}.
Conventional ML, known as centralized learning (CL), relies on aggregating raw data at a central server for model training. 
However, with the exponential growth of global data creation, estimated at 173.4 zettabytes in 2025 and forecasted to triple to 527.5 zettabytes by 2029~\cite{2,lin2023pushing}, CL encounters severe bottlenecks regarding communication overhead and privacy leakage~\cite{2-1,2-2}.
To address these issues, federated learning (FL)~\cite{3,3-1,fang2026hfedmoe} has emerged as a privacy-preserving paradigm, which enables collaborative training by exchanging model updates rather than raw data. 
While FL effectively mitigates data privacy and communication concerns, it confronts severe scalability bottlenecks as modern ML models expand exponentially in size~\cite{4,4-1,lin2026satellite,fang2026aggregation,lin2024fedsn}.

To overcome these bottlenecks,  split learning (SL)~\cite{7,lin2025hasfl} has emerged to partition the model layer-wise, offloading the compute-intensive workload to edge server.
However, vanilla SL adopts sequential training pipeline, which inherently suffers from excessive latency and system overhead due to the idle waiting time on clients~\cite{NSC}. 
To address this limitation, split federated learning (SFL)~\cite{lin2024adaptsfl,6,lin2025hsplitlora} has been proposed by parallelizing client training, yet it reintroduces a significant communication burden by requiring clients to periodically synchronize their local sub-models with a parameter server. Consequently, parallel split learning (PSL)~\cite{8,9} was proposed as a highly streamlined architecture. By completely eliminating client-side parameter aggregation, PSL successfully reduces communication overhead and system complexity, making it the most resource-efficient paradigm for deploying large-scale models on the edge.

However, PSL introduces severe training instability in heterogeneous environments.
In PSL, each client generates intermediate activations based solely on its local, typically non-IID data distribution~\cite{10-1,PipeliningSL}.
In the absence of client-side synchronization, each client's sub-model rapidly overfits to its local data distribution.
This local divergence causes clients to upload semantically inconsistent activations to the server~\cite{11,12}. 
Consequently, during server-side backpropagation, the global model receives wildly divergent and conflicting gradients. 
Rather than converging, the accumulation of these conflicting updates causes destructive interference, leading to severe client drift and potential convergence failure. 
Since reintroducing client-side aggregation would destroy PSL's communication efficiency, the critical challenge is how to resolve these gradient conflicts entirely on the server.

To address the training instability inherent in PSL, we propose PSL-RGHA, a gradient-aligned \underline{PSL} framework that employs a \underline{r}eference \underline{g}radient \underline{h}alf-space \underline{a}lignment mechanism to mitigate conflicting updates among client devices. 
PSL-RGHA comprises two synergistic components designed to enforce global consistency without client-side aggregation. 
Specifically, to approximate the global descent direction, we introduce reference gradient progressive construction (RGPC), which progressively filters and synthesizes directionally consistent client gradients into a robust reference vector. 
Building upon this trusted reference, we devise a gradient half-space alignment (GHA) scheme that projects conflicting client gradients onto the non-negative half-space of the reference direction, effectively neutralizing components that impede convergence.
This collaboration ensures that local updates are continuously harmonized with the global convergence trend without requiring client-side aggregation.
The main contributions of this paper are summarized as follows:
\begin{itemize}[topsep=1pt, itemsep=1pt, parsep=0pt]
   \item We propose PSL-RGHA, a novel gradient alignment framework for PSL that explicitly mitigates gradient conflicts in heterogeneous data environments via pure server-side geometric alignment.
    \item We develop the RGPC algorithm, which utilizes directional consistency metrics and Weiszfeld updates to construct a noise-resilient reference gradient that accurately captures the global convergence trend.
    \item We devise the GHA mechanism, which project client gradients onto the reference half-space, thereby rectifying divergent updates and harmonizing the optimization direction across distributed clients.
    \item We conduct extensive experiments to evaluate PSL-RGHA. Empirical results demonstrate that our framework significantly outperforms existing baselines in both convergence speed and final accuracy across diverse data settings.
\end{itemize}
\section{Related Work}

\subsection{Split Learning and Variants}
SL was originally introduced by Vepakomma~\textit{et al.}~\cite{7} as a privacy-preserving alternative to FL. In vanilla SL framework, the
global model is partitioned into client-side and server-side sub-models. Throughout training, clients interact with the server sequentially in a round-robin fashion to compute forward and backward passes. However, this sequential training incurs severe training latency, hindering scaling  to large numbers of devices. To address the latency bottleneck, Thapa~\textit{et al.}~\cite{6} proposed SFL, which combines SL-style model partitioning with FL-style parallel client training. By allowing multiple clients to perform forward propagation in parallel and aggregating their gradients, SFL significantly reduces the training time compared to vanilla SL. Building upon the parallelization concepts of SFL, recent research has focused on further streamlining the training pipeline. Joshi~\textit{et al.}~\cite{8} and Kim{et al.}~\cite{9} introduced the PSL framework. Unlike SFL, PSL removes the requirement for client-side model aggregation. This architectural change simplifies the synchronization process and accelerates model convergence. Recent work have further improved the computational and communication efficiency of SL by optimizing client-side gradient aggregation, designing scalable server-side update mechanisms, and adopting adaptive model partitioning.

\subsection{Data Heterogeneity Mitigation}

Research on mitigating data heterogeneity in distributed learning has been largely driven by FL, primarily through aggregation optimization and personalization. Representative methods such as SCAFFOLD~\cite{13}, FedDyn~\cite{14}, and FedDC~\cite{15} alleviate client drift via control variates or regularization, while personalization approaches (e.g., FedAPA~\cite{17}, FedCPD~\cite{18}) enhance adaptability under non-IID data through adaptive aggregation or prototype-based learning.
More recently, several studies have addressed data heterogeneity through update consistency and conflict-aware optimization. Methods such as FedProx~\cite{FedProx}, FedNova~\cite{FedNova}, and MOON~\cite{moon} improve training stability by constraining local drift, correcting objective inconsistency, or enhancing representation consistency. Conflict-aware approaches, including PCGrad~\cite{pcgrad}, FedIIR~\cite{FedIIR}, FedLD~\cite{FedLD}, and FedCALM~\cite{FedCALM}, further mitigate interference among client updates by aligning optimization directions or suppressing conflicting components.
However, these methods are mainly designed for FL, where full-model replication and iterative client aggregation are assumed. By contrast, PSL adopts a partitioned architecture without client-side aggregation, leading to distinct gradient interaction patterns. Therefore, mitigating gradient inconsistency in PSL requires methods tailored to its unique training dynamics.
\section{Background and Motivation}

PSL has emerged as a promising training paradigm for resource-constrained edge intelligence.
By partitioning the model between edge devices and the server, PSL offloads the majority of computation and memory overhead from clients. Crucially, by completely eliminating client-side parameter aggregation, PSL simplifies the training pipeline and minimizes communication overhead, making it highly attractive for large-scale distributed deployment compared to conventional SFL.
However, this highly efficient, client-side parameter aggregation-free architecture introduces an inherent optimization vulnerability under statistical heterogeneity. 
In PSL, each client executes forward and backward propagation based solely on its local, typically non-IID data, updating its sub-model without any cross-client synchronization. 
Consequently, local sub-models rapidly overfit to local distributions. 
This local bias causes clients to upload semantically disjointed intermediate activations to the server. 
Under non-IID settings, this semantic mismatch translates into highly divergent gradients during server-side backpropagation. 
Instead of a globally calibrated optimization signal, the server receives conflicting updates that accumulate over time, leading to optimization divergence and severe client drift.
\begin{figure}[!t]
    \captionsetup[subfloat]{skip=-0pt}
    \centering
    \begin{subfigure}[b]{0.42\linewidth}
        \vspace{0pt}
        \centering
        \includegraphics[width=\linewidth]{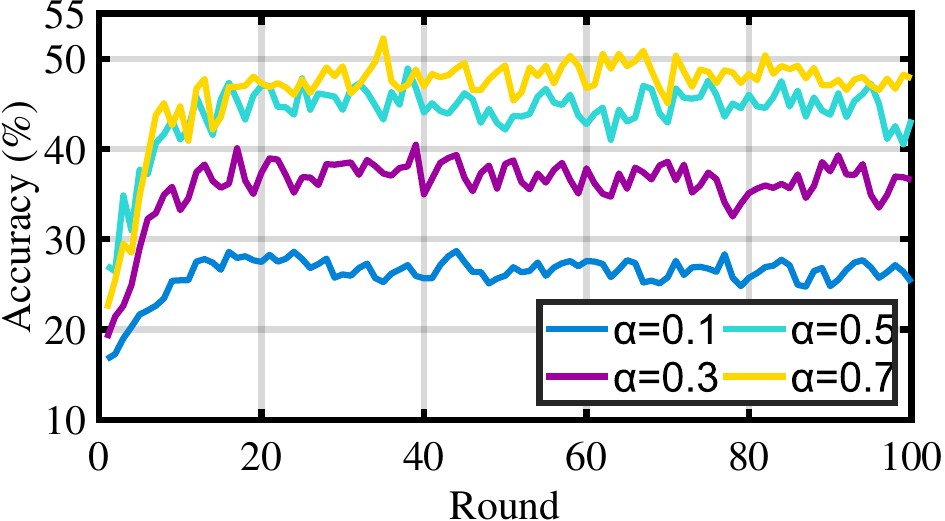} 
   
        \caption{PSL}
    \end{subfigure}
    \hspace{1em}
    \begin{subfigure}[b]{0.42\linewidth}
        \vspace{0pt}
        \centering
        \includegraphics[width=\linewidth]{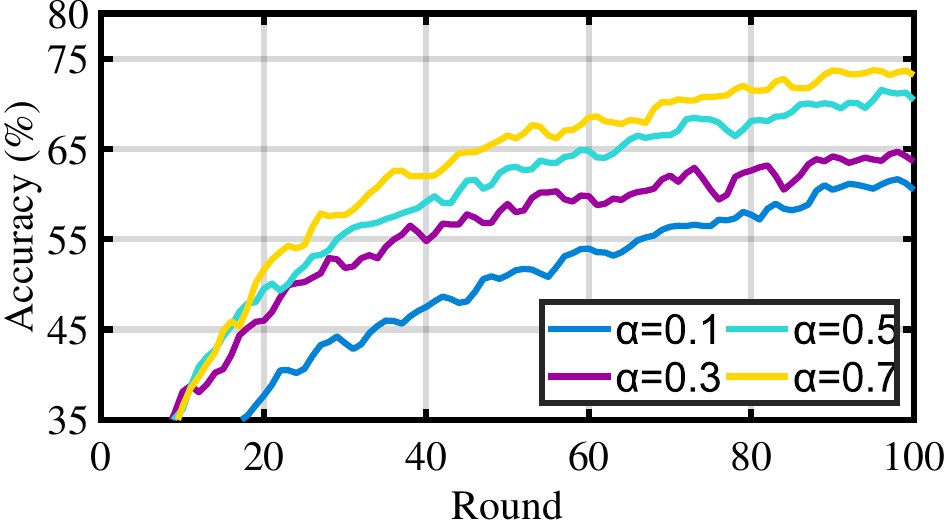}
      
        \caption{SFL }
    \end{subfigure}

    \caption{Performance of PSL and SFL on CIFAR-10 datasets under non-IID settings with ResNet-18.}
    \label{fig:yushiyan1}
    \vspace{0pt}
\end{figure}

\begin{figure}[t]
    \captionsetup[subfloat]{skip=0pt}
    \centering
    \begin{subfigure}[b]{0.42\linewidth}
        \vspace{0pt}
        \centering
        \includegraphics[width=\linewidth]{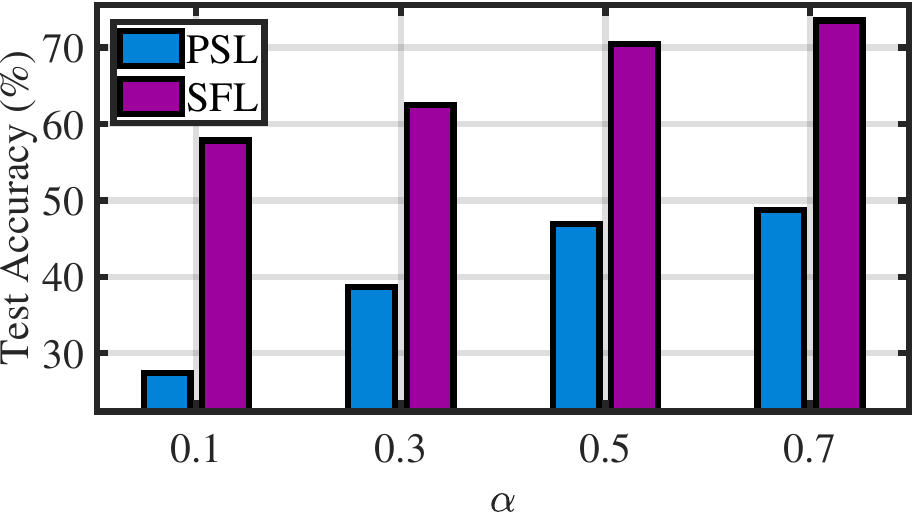} 
   
        \caption{Test accuracy}
        \label{fig:yu1}
    \end{subfigure}
    \hspace{1em}
    \begin{subfigure}[b]{0.42\linewidth}
        \vspace{0pt}
        \centering
        \includegraphics[width=\linewidth]{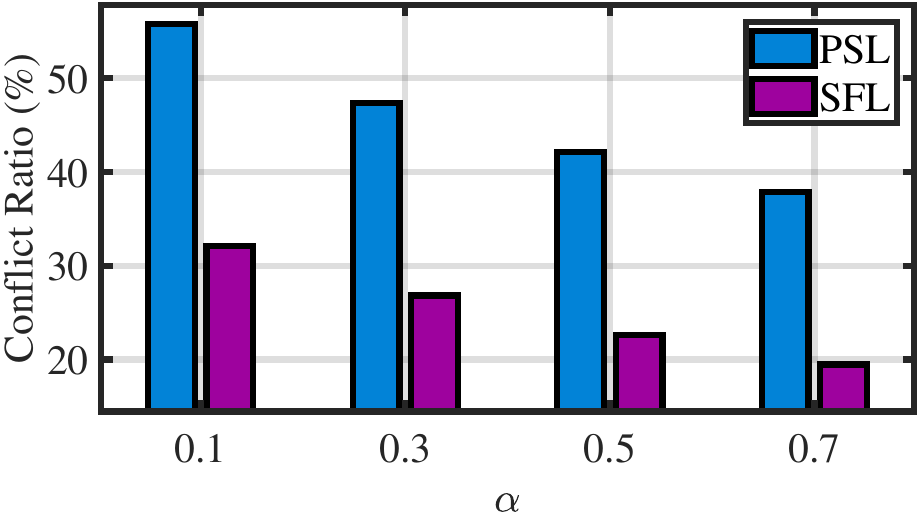}
      
        \caption{Conflict ratio }
        \label{fig:yu2}
    \end{subfigure}

    \caption{Gradient conflict ratio and accuracy of PSL and SFL under different non-IID levels on CIFAR-10 with ResNet-18.}
    \vspace{-0pt}
    \label{fig:yushiyan2}
\end{figure}

This vulnerability is supported by our preliminary experiments. As shown in Fig.~\ref{fig:yushiyan1}, on non-IID CIFAR-10 with ResNet-18, PSL converges more slowly and less stably than SFL, suggesting that the absence of client-side aggregation increases its sensitivity to data heterogeneity. Fig.~\ref{fig:yushiyan2} further shows that stronger heterogeneity leads to a higher gradient conflict ratio and lower test accuracy, with a more pronounced degradation in PSL. These results indicate that PSL's performance loss is closely associated with cross-client gradient inconsistency, which introduces noisy and unstable server-side updates.
The above observations reveal a major limitation of existing PSL frameworks: eliminating client-side aggregation maximizes communication efficiency but simultaneously removes the essential cross-client calibration needed for stable convergence.  
Consequently, heterogeneous client updates can interfere with each other, causing oscillation, slow convergence, or even convergence failure. Therefore, the key challenge is how to preserve the efficiency advantage of aggregation-free PSL while still enforcing directional consistency among heterogeneous client gradients. 
This compelling dilemma motivates the necessity of our research: designing a robust server-side coordination mechanism capable of inferring a reliable global descent trend from noisy client signals, and actively suppressing conflicting update directions to guarantee stable, aggregation-free distributed training.

\begin{figure}[t]
    \centering
    \includegraphics[width=0.45\textwidth]{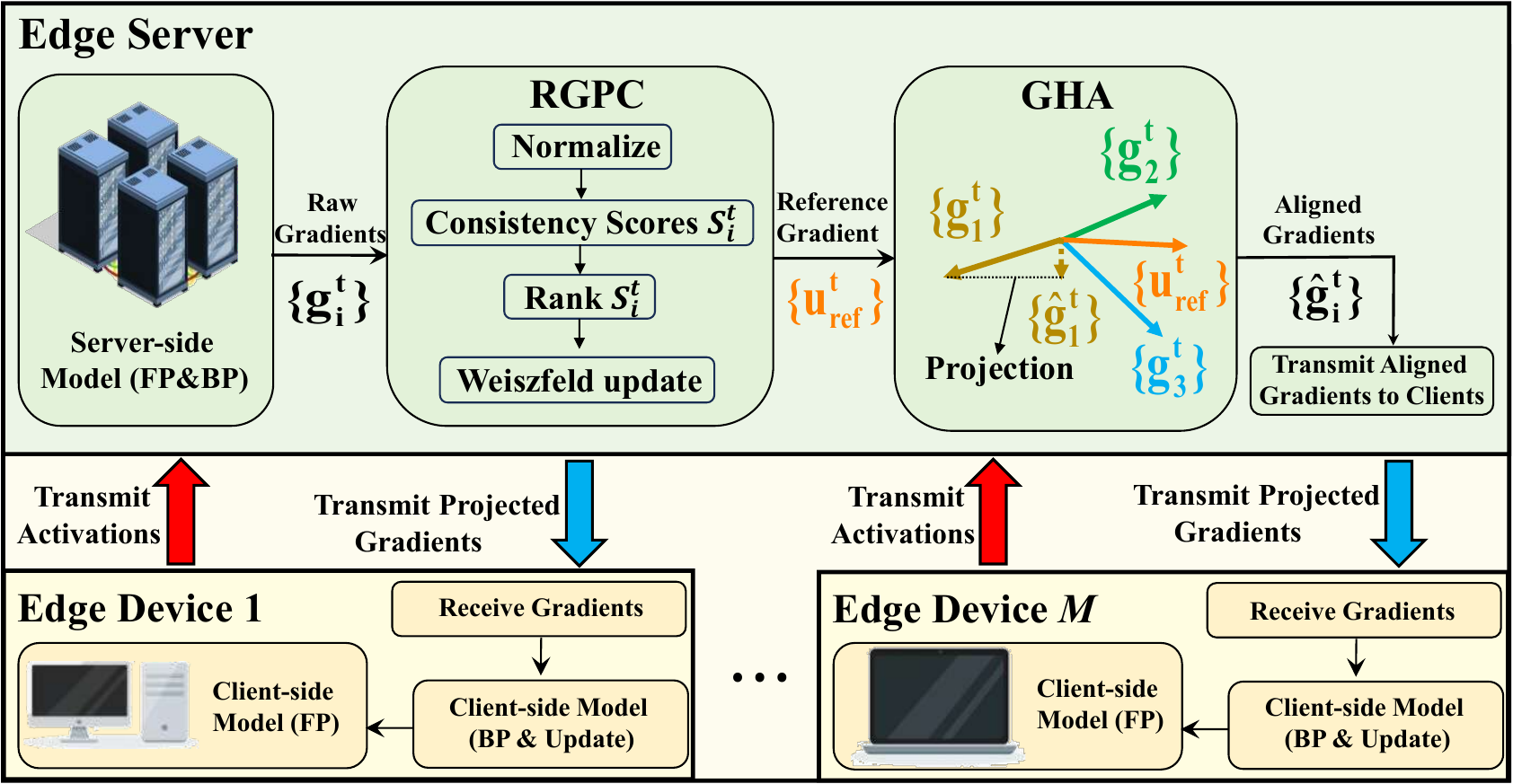} 
    \vspace{2pt}
    \caption{An overview of PSL-RGHA architecture.}
    \label{fig:1} 
    \vspace{0pt}
\end{figure}

\section{System Design}
\subsection{Overview}

In this section, we present PSL-RGHA, a robust gradient-alignment framework engineered to stabilize PSL under severe statistical heterogeneity. 
As illustrated in Fig.~\ref{fig:1}, PSL-RGHA integrates two synergistic modules within an aggregation-free architecture: i) RGPC (Section~\ref{ssec:RGAC}), which synthesizes a global reference direction by progressively aggregating consistent client gradients via Weiszfeld updates; ii)GHA (Section ~\ref{ssec:GHA}), which leverages the reference gradient to enforce global directional consistency. 

\begin{figure}[t]
    \captionsetup[subfloat]{skip=-0.5pt}
    \centering
    \begin{subfigure}[b]{0.41\linewidth}
        \vspace{0pt}
        \centering
        \includegraphics[width=\linewidth]{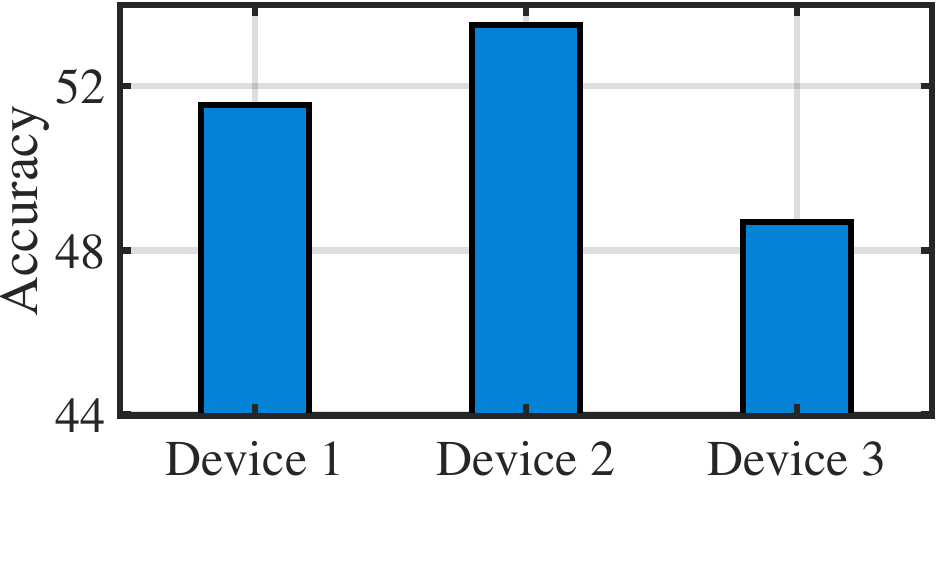} 
        \caption{Individual performance}
        \label{fig3a}
        \vspace{1pt}
    \end{subfigure}
    \hspace{1em}
    \begin{subfigure}[b]{0.42\linewidth}
       \vspace{0pt}
        \centering
        \includegraphics[width=\linewidth]{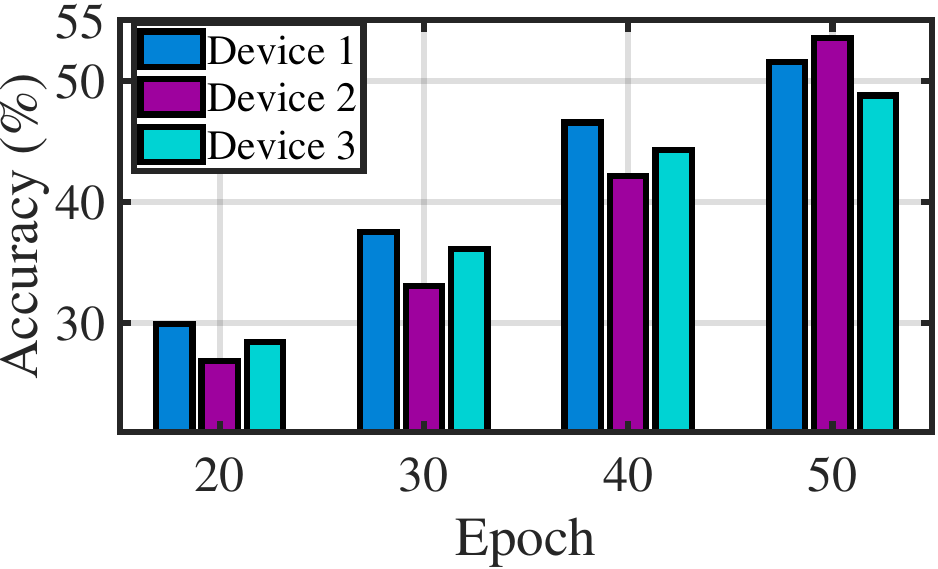}
        \caption{Accuracy vs. Epochs}
        \label{fig3b}
        \vspace{1pt}
    \end{subfigure}
    \caption{Performance of PSL using single client device gradient updates on CIFAR-10 under non-IID settings (ResNet-18).}
    \label{fig:3}
\end{figure}

In PSL-RGHA, the model is split into client-side sub-models and a shared server-side sub-model. In each training round, clients perform forward propagation on local data and upload cut-layer activations to the server. The server then completes forward and backward propagation to obtain client-specific server gradients. RGPC constructs a robust reference gradient, and GHA aligns each client gradient within the corresponding reference half-space. The server updates the shared sub-model using the corrected gradients and returns the cut-layer gradients to the clients, which complete backpropagation and update their local sub-models.

\subsection{\hspace{-3pt}Reference Gradient Progressive Construction}
\label{ssec:RGAC}

In PSL, the absence of client-side aggregation severely exacerbates inter-client gradient misalignment, producing conflicting updates at the server. To empirically validate this, we evaluated the optimization trajectories of individual clients starting from identical initializations. As illustrated in Fig.~\ref{fig3a}, the resultant test accuracies exhibit extreme spatial variance, confirming the severe heterogeneity of local gradients. Furthermore, the client contributing the maximum performance gain fluctuates across epochs (Fig.~\ref{fig3b}), indicating that no single client consistently aligns with the optimal global descent direction. Consequently, under such directional conflict, naive uniform aggregation inadvertently incorporates these misaligned gradients as detrimental noise, thereby degrading the global update and stalling convergence.

\begin{figure}[t]
    \vspace{0pt}
    \centering
    \includegraphics[width=0.45\textwidth]{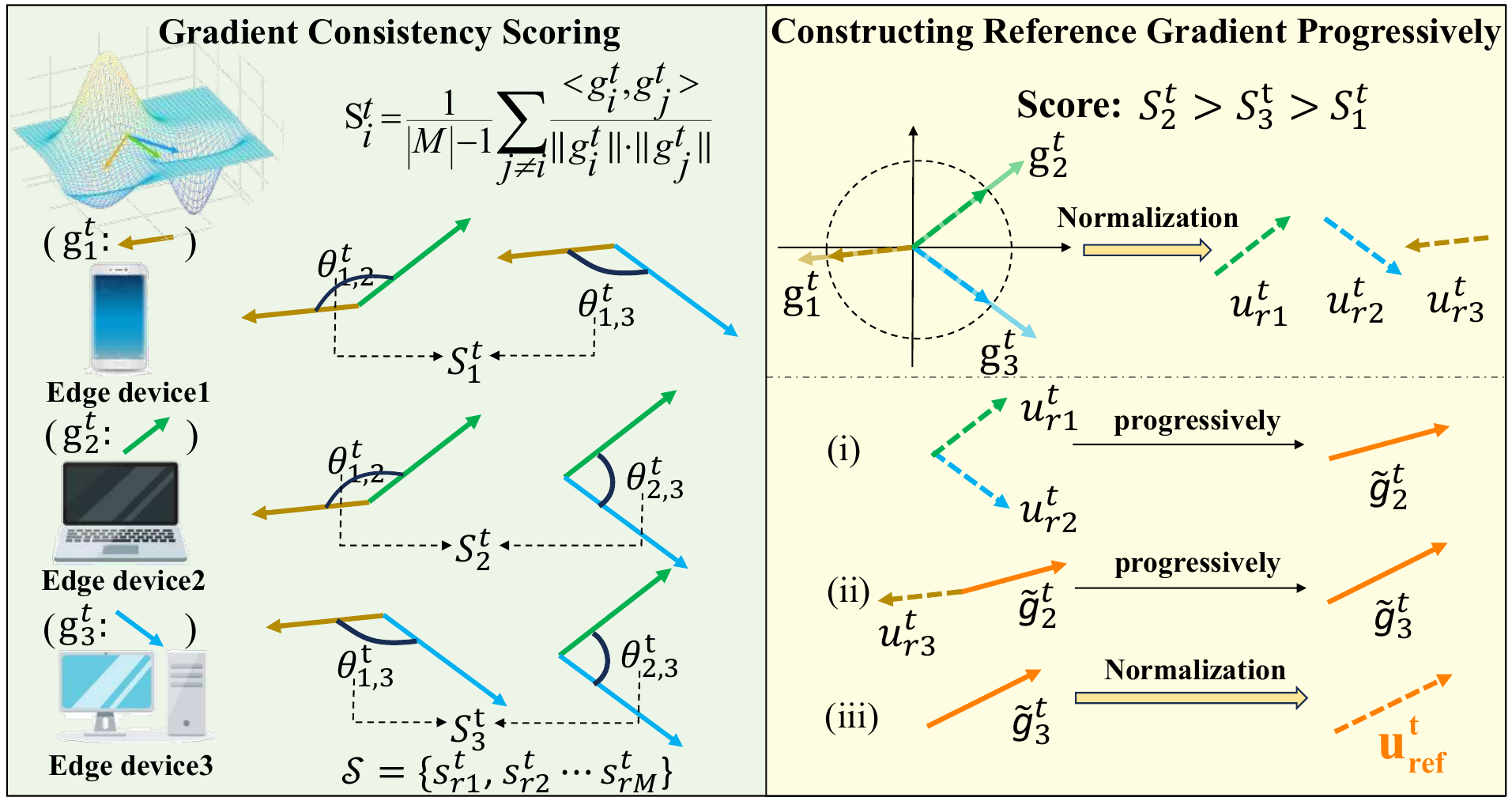} 
    \caption{An illustration of RGPC: Gradients are scored by directional consistency (left) and aggregated via ranked Weiszfeld updates (right) to construct the reference gradient.} 
    \label{fig:2} 
\end{figure}

To this end, we propose RGPC, a robust aggregation mechanism that ranks client gradients via a directional consistency score and progressively synthesizes them using Weiszfeld updates~\cite{25-1} , as illustrated in Fig.~\ref{fig:2}. 
This procedure yields a robust reference gradient that effectively filters out noise and captures the underlying global convergence trend, which is elaborated as follows.

\subsubsection{Gradient Consistency Scoring}
Existing methods often estimate client contributions using gradient magnitudes, assuming that larger updates are more informative~\cite{24,24-1,24-2}. 
However, under statistical heterogeneity (non-IID data), gradients with large magnitudes often stem from local overfitting or outliers, which are directionally inconsistent with the global optimization landscape and may impede convergence. 
Conversely, gradients with smaller magnitudes but consistent directions often provide reliable progress.
This necessitates a metric that prioritizes directional alignment over scalar magnitude. Accordingly, we introduce the \textit{gradient consistency score}, a metric designed to quantify the cross-client directional agreement. Formally, for the $t$-th training round, the consistency score $S_{i}^{t}$ of the $i$-th client is defined as the average cosine similarity between its local gradient and those of all other participating clients:
\vspace{-5pt}
\begin{equation}
S_i^{t}
=\frac{1}{|\mathcal{M}|-1}\sum_{j\neq i}
\frac{\left\langle g_i^{t},\, g_j^{t}\right\rangle}
{\left\lVert g_i^{t}\right\rVert \cdot \left\lVert g_j^{t}\right\rVert } .
\label{eq:1}
\vspace{-2pt}
\end{equation}  where $\mathcal{M}=\{1,2,\ldots,M\}$ denotes the set of participating client devices,
⟨·, ·⟩ denotes the inner product, $\lVert \cdot \rVert$ is the Euclidean norm, and $g_{k}^{t}$ is the gradient of the $k$-th client device at the $t$-th training round.

\begin{figure}[t]
    \captionsetup[subfloat]{skip=0.5pt}
    \centering
    \begin{subfigure}[b]{0.42\linewidth}
        \vspace{0pt}
        \centering
        \includegraphics[width=\linewidth]{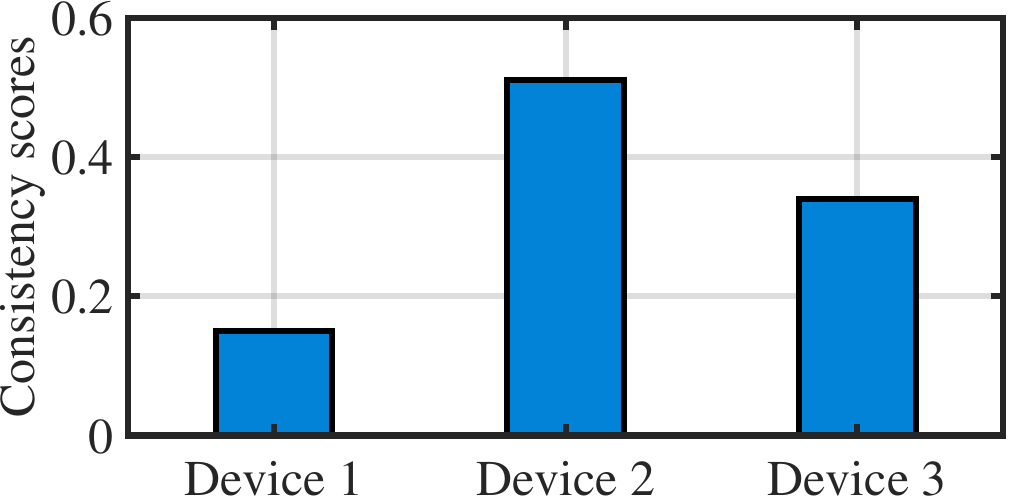} 
   
        \caption{Consistency scores}
        \label{fig:4a}
    \end{subfigure}
    \hspace{1em}
    \begin{subfigure}[b]{0.42\linewidth}
        \vspace{0pt}
        \centering
        \includegraphics[width=\linewidth]{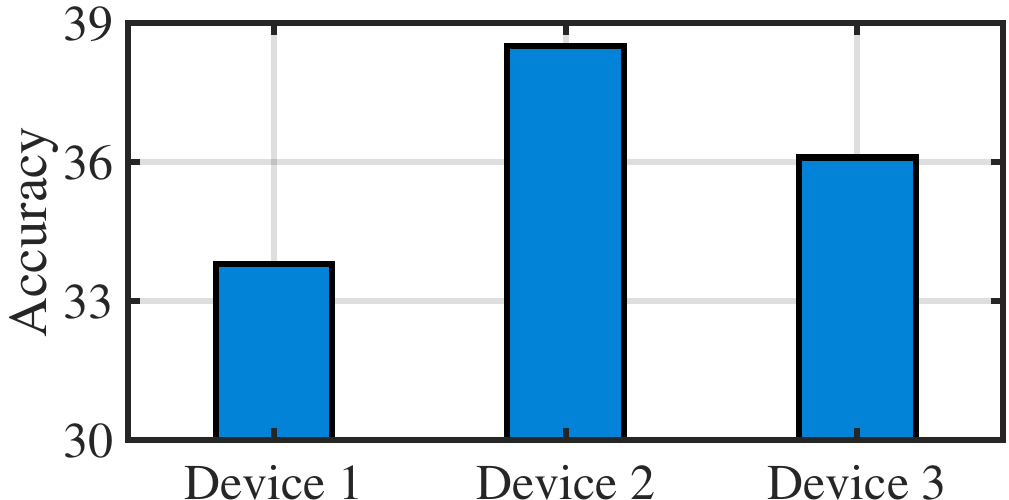}
      
        \caption{Accuracy }
        \label{fig:4b}
    \end{subfigure}
    
  \vspace{0pt}
    \caption{The gradient consistency scores and test accuracies of client devices on CIFAR-10 (ResNet-18).}
    \label{fig:4}
    \vspace{0pt}
\end{figure}

Equ.~(\ref{eq:1}) provides a quantitative measure of directional centrality. A higher $S_i^t$ indicates that the gradient of client $i$ aligns closely with the prevailing descent trend of the population, suggesting a constructive contribution to the global model. 
Conversely, a lower $S_i^t$ signifies weak directional agreement, implying that the update may introduce noise or conflicting signals that decelerate convergence. 
As shown in Fig.~\ref{fig:4}, clients with higher consistency scores consistently yield higher test accuracy in single-client update studies, validating $S_i^{t}$ as an effective indicator of optimization utility.

After computing $\{S_i^t\}_{i\in\mathcal{M}}$, the server ranks clients in descending order and obtains indices $(r_1, r_2 \ldots,r_M)$ such that
 \vspace{-3pt}
\begin{equation}
S_{r_1}^{t} \ge S_{r_2}^{t} \ge \cdots \ge S_{r_M}^{t},
\label{eq:rank}
\end{equation}
where $r_1$ denotes the client with the highest directional consistency at round $t$. This ranking serves as the basis for RGPC, allowing the system to prioritize directionally reliable gradients while mitigating the impact of outliers during the subsequent reference construction.



\subsubsection{Progressive Construction via Weiszfeld Updates}
To synthesize a robust reference gradient from the heterogeneous client updates, we employ a progressive Weiszfeld~\cite{25-2} scheme that prioritizes directionally consistent gradients (i.e., those with higher consistency scores $S_i^t$). Unlike simple averaging which is sensitive to outliers, this approach sequentially aggregates ranked directions to approximate the geometric median, which is robust especially in high-dimensional spaces.

Given the set of client gradients  $\{g_i^{t}\}_{i\in\mathcal{M}}$ at round $t$, we first decouple directional information from magnitude by normalizing valid gradients onto the unit sphere:
\vspace{-2pt}
\begin{equation} {u}_i^{t}=\frac{{g}_i^{t}}{\| {g}_i^{t}\|},\quad \forall i\in\mathcal{M}, \text{ s.t. } \|{g}_i^{t}\| > \delta, \label{eq:unit_dir} 
\end{equation}
\vspace{-2pt}
where $\delta$ is a threshold to filter numerically unstable updates.

Let $(r_1,r_2\ldots,r_M)$ denote the indices of clients sorted by their consistency scores in descending order. 
We process clients according to this rank, maintaining an evolving estimate $\boldsymbol{g} \in \mathbb{R}^d$ constrained within the convex unit ball $\mathcal{C}\triangleq\{\boldsymbol{x}\in\mathbb{R}^d:\|\boldsymbol{x}\|\le 1\}$. Crucially, we perform intermediate updates within $\mathcal{C}$ (a convex set) rather than the unit sphere (non-convex) to guarantee stable convergence properties.
For the cumulative subset $\mathcal{P}_k^{t}=\{u_{r_1}^{t},\ldots,u_{r_k}^{t}\}$ of the top-$k$ clients, the smoothed geometric-median objective over $\mathcal{C}$ is defined as:
\begin{equation}
\Phi_{k,\varepsilon}^{t}(g)=\sum_{\ell=1}^{k}\sqrt{\|g-u_{r_\ell}^{t}\|^{2}+\varepsilon^{2}},\quad g\in\mathcal{C}.
\label{eq:gm_obj}
\end{equation}
where $\ell\in\{1,2,\ldots,k\}$ is the descent index of clients.

We initialize the estimate as ${g} \leftarrow {u}_{r_1}^{t}$ and sequentially incorporate the ranked directions. Upon adding the $k$-th direction, we perform one warm-started Weiszfeld step using the current estimate ${g}$ (obtained from the previous $k-1$ steps) to update the centroid:
\vspace{0pt}
\begin{subequations}
\begin{align}
w_\ell^{(k)} &= \frac{1}{\sqrt{\|g-u_{r_\ell}^{t}\|^{2}+\varepsilon^{2}}},\quad  \label{eq:weisz_w}\\
g &\leftarrow \frac{\sum_{\ell=1}^{k} w_\ell^{(k)}\,u_{r_\ell}^{t}}{\sum_{\ell=1}^{k} w_\ell^{(k)}}, \label{eq:weisz_avg}
\end{align}
\end{subequations}
where $\varepsilon>0$ ensures numerical stability. 
After processing all ranked gradients, we obtain the reference direction by normalizing the final estimate:
\vspace{0pt}
\begin{equation}
u_{\mathrm{ref}}^{t} = \frac{ g}{\| g\|} \quad (\|g\| > 0).
\label{eq:uref_final}
\vspace{0pt}
\end{equation}
This progressive construction strategy confers robustness through a dual mechanism: (i) the consistency-based ranking ensures that the reference vector is anchored by the most reliable gradients during the critical early phases, thereby mitigating initialization bias; and (ii) the Weiszfeld reweighting naturally attenuates the influence of updates with large angular deviations (outliers), effectively synthesizing a noise-resilient global descent direction. 
Subsequently, the server broadcasts ${u}_{\mathrm{ref}}^{t}$ to all participating clients to orchestrate the gradient alignment in the forthcoming GHA phase.




\subsection{Gradient Half-space Alignment}
\label{ssec:GHA}
\begin{figure}[t]
     \vspace{0pt}
    \centering
    \includegraphics[width=0.4\textwidth]{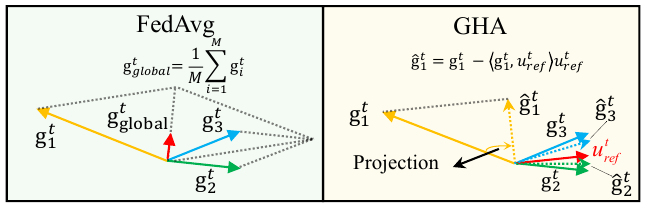} 
    \caption{The gradient alignment result of FedAvg and GHA.}
    \label{fig:5} 
    \vspace{0pt}
\end{figure}

The reference gradient $\boldsymbol{u}_{\mathrm{ref}}^{t}$ constructed by RGPC serves as a robust proxy for the global optimization trajectory. 
In conventional aggregation schemes such as FedAvg, the server indiscriminately averages client updates, neglecting their directional disparities. 
As shown in Fig.~\ref{fig:5}, this simple aggregation allows conflicting gradients to destructively interfere, leading to optimization drift or cancellation in the global model.

To mitigate this, we propose the GHA mechanism, which enforces directional agreement by geometrically projecting conflicting gradients onto the reference half-space and softly aligning non-opposing updates toward the reference direction.
Therefore, GHA effectively neutralizes divergent components while preserving the informative orthogonal innovations.
Formally, given the server-side reference direction $\boldsymbol{u}_{\mathrm{ref}}^{t}$ at round $t$, we define the permissible reference half-space:
\[
\mathcal{H}^{t}\triangleq\{d\in\mathbb{R}^d:\langle d,u_{\mathrm{ref}}^{t}\rangle\ge 0\},
\]
which contains all directions that form a non-obtuse angle with $u_{\mathrm{ref}}^{t}$.
For each client update $g_i^{t}$, GHA applies a piecewise alignment rule:
\begin{align}
\hat g_i^{t}=
\begin{cases}
(1-\lambda_i^{t})\,g_i^{t}+\lambda_i^{t}\,\|g_i^{t}\|\,u_{\mathrm{ref}}^{t},
& \langle g_i^{t},u_{\mathrm{ref}}^{t}\rangle\ge 0,\\[0.1em]
g_i^{t}-\langle g_i^{t},u_{\mathrm{ref}}^{t}\rangle\,u_{\mathrm{ref}}^{t},
& \langle g_i^{t},u_{\mathrm{ref}}^{t}\rangle<0,
\end{cases}
\label{eq:gha_projection}
\end{align}
where softly aligns non-opposing updates toward the reference ray (while preserving the update norm), and performs the Euclidean projection onto $\mathcal{H}^{t}$ to remove only the opposing component along $u_{\mathrm{ref}}^{t}$ while preserving orthogonal information.


Geometrically, Eq.~\eqref{eq:gha_projection} aligns each client update with the reference half-space. For non-conflicting updates (Case 1), GHA softly pulls $g_i^{t}$ toward the reference ray $||g_i^{t}||u_{\mathrm{ref}}^{t}$, with smaller perturbations for more consistent updates. For conflicting updates (Case 2), it removes only the negative component along $u_{\mathrm{ref}}^{t}$ while preserving the orthogonal component. Thus, every corrected update satisfies $\langle \hat g_i^{t},u_{\mathrm{ref}}^{t}\rangle\ge 0$, supporting the global descent direction without discarding client-specific information. GHA requires only $O(d)$ vector operations and introduces no additional communication. The full procedure is given in Algorithm~\ref{alg:psl_RGHA}.

\begin{algorithm}[t]
\caption{PSL-RGHA}
\small
\label{alg:psl_RGHA}
\begin{algorithmic}[1]
\STATE \textbf{Input:} Client gradients $\{g_i^{t}\}_{i\in\mathcal{M}}$, smoothing $\varepsilon>0$
\STATE \textbf{Output:} Reference gradient $u_{\mathrm{ref}}^{t}$, corrected updates $\{\hat g_i^{t}\}_{i\in\mathcal{M}}$

\vspace{0.2em}
\STATE \textbf{Phase 1: Consistency scoring and ranking.}
\FOR{each $i\in\mathcal{M}$}
    \STATE $S_i^{t}\leftarrow \frac{1}{|\mathcal{M}|-1}\sum_{j\in\mathcal{M},\,j\neq i}
    \frac{\langle g_i^{t},g_j^{t}\rangle}{\|g_i^{t}\|\cdot\|g_j^{t}\|}$
\ENDFOR
\STATE Sort to obtain $(r_1,\ldots,r_M)$ such that $S_{r_1}^{t}\ge\cdots\ge S_{r_M}^{t}$

\vspace{0.2em}
\STATE \textbf{Phase 2: Progressive construction via Weiszfeld updates.}
\STATE $g \leftarrow u_{r_1}^{t}$ \COMMENT{$u_i^{t}\triangleq g_i^{t}/\|g_i^{t}\|$}
\FOR{$k=2$ \textbf{to} $M$}
    \STATE $w_\ell^{(k)} \leftarrow \big(\|g-u_{r_\ell}^{t}\|^2+\varepsilon^2\big)^{-1/2}$ for $\ell=1,\ldots,k$
    \STATE $ g \leftarrow \frac{\sum_{\ell=1}^{k} w_\ell^{(k)}\,u_{r_\ell}^{t}}{\sum_{\ell=1}^{k} w_\ell^{(k)}}$
\ENDFOR
\STATE $u_{\mathrm{ref}}^{t} \leftarrow  g/\| g\|$
\vspace{0.2em}
\STATE \textbf{Phase 3: Gradient half-space alignment.}
\FOR{each client $i\in\mathcal{M}$ \textbf{in parallel}}
    \STATE $s \leftarrow \langle g_i^{t}, u_{\mathrm{ref}}^{t}\rangle$
    \IF{$s \ge 0$}
    \STATE $\lambda \leftarrow \frac{1}{2}\left(1-\frac{s}{\|g_i^{t}\|}\right)$
    \STATE $\hat g_i^{t} \leftarrow (1-\lambda)\,g_i^{t}+\lambda\,\|g_i^{t}\|\,u_{\mathrm{ref}}^{t}$

    \ELSE
        \STATE $\hat g_i^{t} \leftarrow g_i^{t} - s\cdot u_{\mathrm{ref}}^{t}$
    \ENDIF
\ENDFOR

\STATE \textbf{return} $u_{\mathrm{ref}}^{t}, \{\hat g_i^{t}\}_{i\in\mathcal{M}.}$
\end{algorithmic}
\end{algorithm}

\subsection{Theoretical Analysis}

We provide theoretical insights for the two key modules of PSL-RGHA. For RGPC, we demonstrate that progressively incorporating ranked client directions---and performing a warm-started Weiszfeld refinement after each step---follows a Majorization-Minimization (MM) framework over the convex set $\mathcal{C}$. Consequently, the smoothed geometric-median objective $\Phi_{k,\varepsilon}^{t}(\cdot)$ monotonically decreases after each refinement. This guarantees that RGPC reliably constructs a robust reference gradient direction $u_{\mathrm{ref}}^{t}$ to proxy the global descent trend. Building on this reference direction, we further analyze the geometric properties and stability of GHA. GHA projects conflicting updates onto the reference half-space and softly aligns non-conflicting ones toward $u_{\mathrm{ref}}^{t}$. Theorem~\ref{thm:gha_geom_compact} proves that these corrected updates strictly reside in the valid half-space, improve global alignment, and remain non-expansive in norm, which fundamentally ensures training stability under heterogeneous gradient distributions.

\begin{theorem}[MM descent of a progressive Weiszfeld step]
\label{thm:mm_weiszfeld}
Fix a training round $t$ and a prefix size $k$. Let the feasible set be $\mathcal{C}\triangleq\{g\in\mathbb{R}^d:\|g\|\le 1\}$ and the smoothing parameter be $\varepsilon>0$. We define the smoothed geometric-median objective as:
\vspace{-3pt}
\begin{equation}
\Phi_{k,\varepsilon}^{t}(g)\triangleq \sum_{\ell=1}^{k}\sqrt{\|g-u_{r_\ell}^{t}\|^{2}+\varepsilon^{2}}, \qquad g\in\mathcal{C}.
\label{eq:phi_smooth}
\end{equation}
Given a current estimate $g^{(0)}\in\mathcal{C}$, we define the distance and corresponding Weiszfeld weights for $\ell=1,\ldots,k$ as:
\begin{equation}
d_\ell \triangleq \sqrt{\|g^{(0)}-u_{r_\ell}^{t}\|^{2}+\varepsilon^{2}}, \quad w_\ell \triangleq \frac{1}{d_\ell}.
\label{eq:weights_smooth}
\end{equation}
Applying a single Weiszfeld update step yields:
\begin{equation}
g^{(1)} \triangleq \frac{\sum_{\ell=1}^{k} w_\ell\,u_{r_\ell}^{t}}{\sum_{\ell=1}^{k} w_\ell}.
\label{eq:update_smooth}
\end{equation}
Then, the updated estimate remains strictly within the convex set ($g^{(1)}\in\mathcal{C}$), and the objective function monotonically decreases:
\begin{equation}
\Phi_{k,\varepsilon}^{t}(g^{(1)})\le \Phi_{k,\varepsilon}^{t}(g^{(0)}).
\label{eq:mono_descent}
\end{equation}
\end{theorem}

{
\begin{figure*}[t]
    \captionsetup[subfloat]{skip=-2pt}
     \vspace{-3pt}
    \centering
    \subfloat[{\fontsize{5.5pt}{6pt}\selectfont ResNet18 CIFAR10 (IID).}]{
        \includegraphics[width=0.16\linewidth]{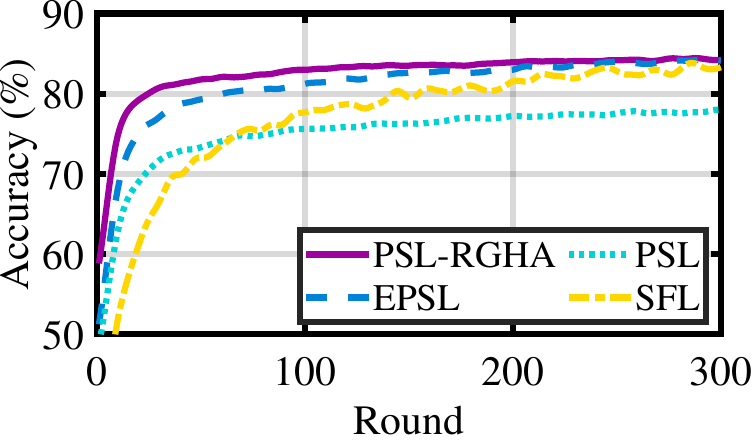}}
    \subfloat[{\fontsize{5.5pt}{6pt}\selectfont ResNet18 CIFAR100 (IID).}]{
        \includegraphics[width=0.16\linewidth]{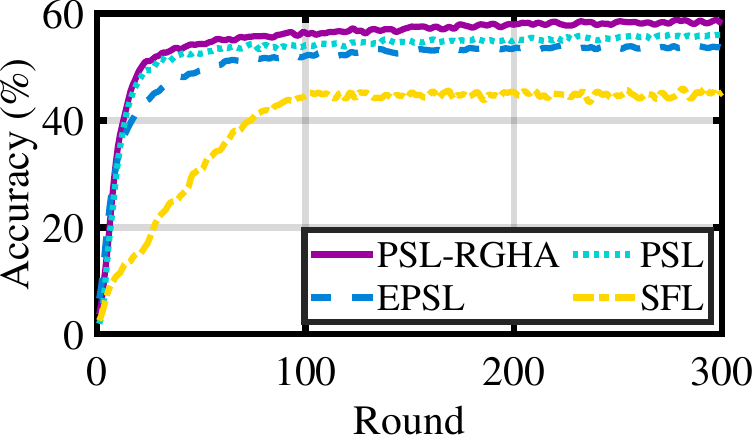}}
    \subfloat[{\fontsize{5.5pt}{6pt}\selectfont ResNet18 FMNIST (IID).}]{
        \includegraphics[width=0.16\linewidth]{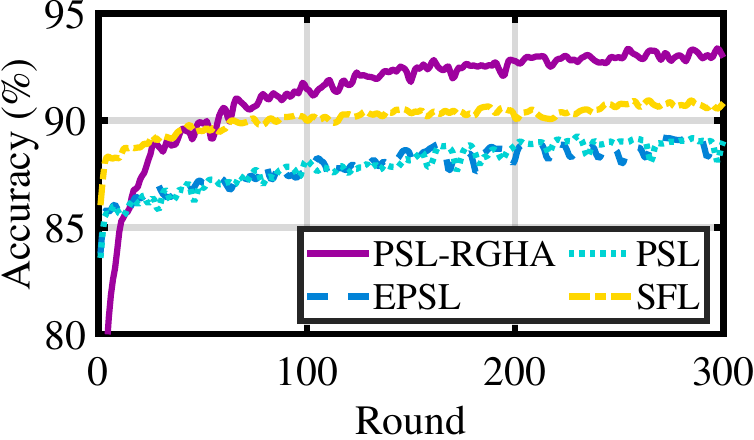}}
    \subfloat[{\fontsize{5.5pt}{6pt}\selectfont ResNet-32 CIFAR10 (IID).}]{
        \includegraphics[width=0.16\linewidth]{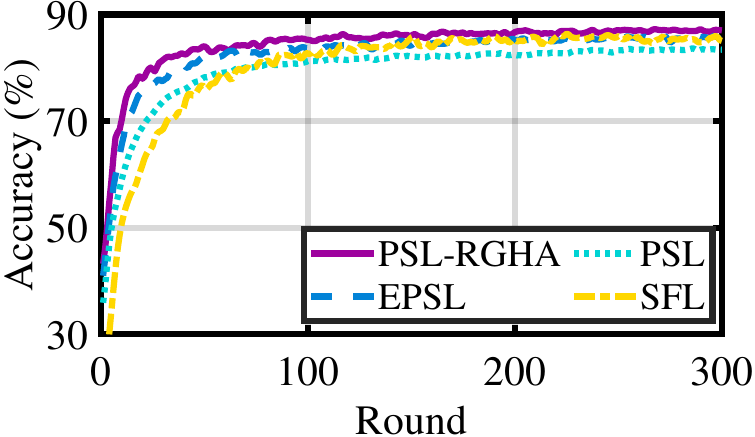}}
    \subfloat[{\fontsize{5.5pt}{6pt}\selectfont  ResNet32 CIFAR100 (IID).}]{
        \includegraphics[width=0.16\linewidth]{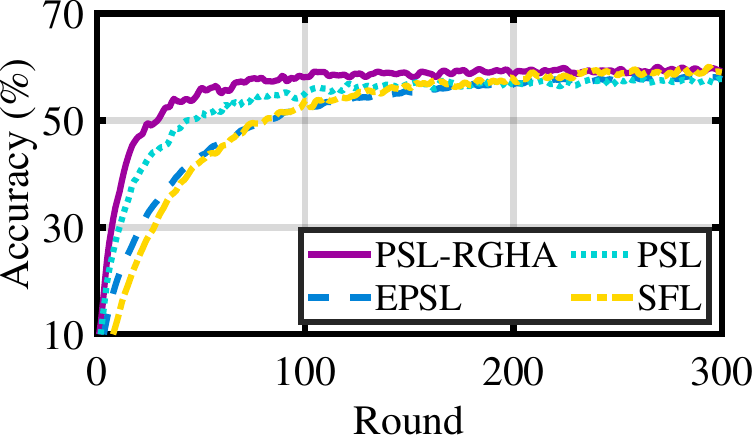}}
    \subfloat[{\fontsize{5.5pt}{6pt}\selectfont  ResNet32 FMNIST (IID).}]{
        \includegraphics[width=0.16\linewidth]{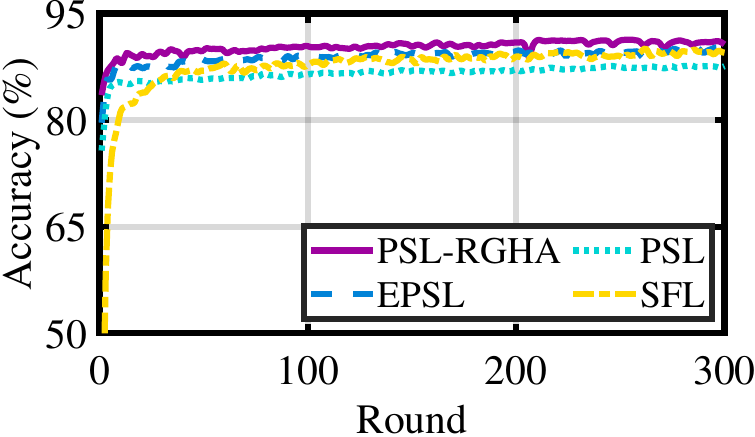}}
    \\[0.5ex]

    \subfloat[{\fontsize{5.5pt}{6pt}\selectfont ResNet18 CIFAR10 (non-IID).}]{
        \includegraphics[width=0.16\linewidth]{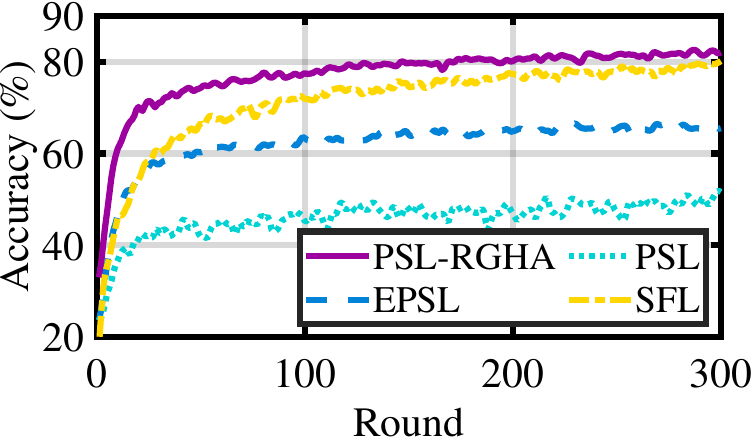}}
    \subfloat[{\fontsize{5.5pt}{6pt}\selectfont ResNet18CIFAR100(non-IID).}]{
        \includegraphics[width=0.16\linewidth]{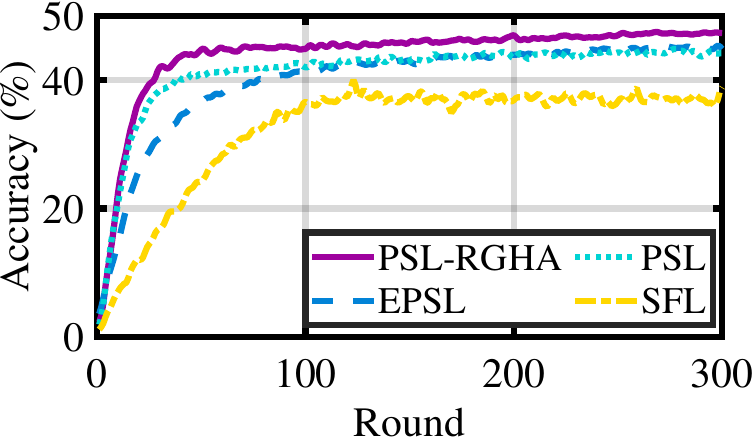}}
    \subfloat[{\fontsize{5.5pt}{6pt}\selectfont ResNet18 FMNIST (non-IID).}]{
        \includegraphics[width=0.16\linewidth]{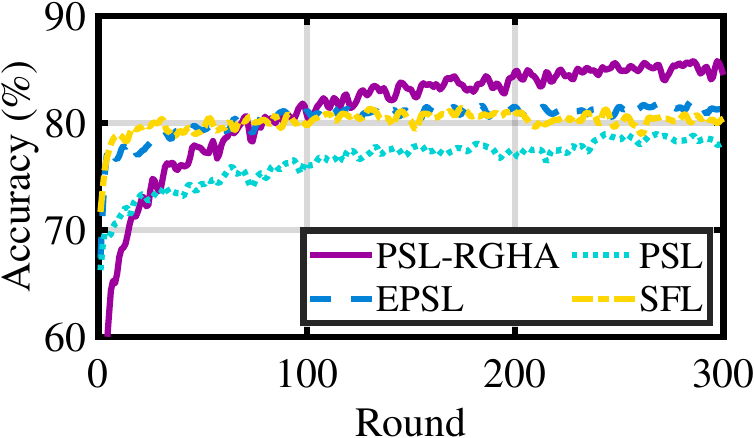}}
    \subfloat[{\fontsize{5.5pt}{6pt}\selectfont ResNet32 CIFAR10 (non-IID).}]{
        \includegraphics[width=0.16\linewidth]{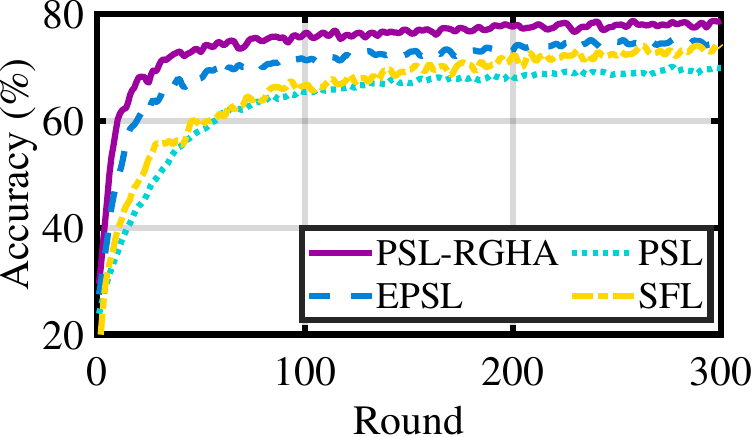}}
    \subfloat[{\fontsize{5.5pt}{6pt}\selectfont ResNet32CIFAR100(non-IID).}]{
        \includegraphics[width=0.16\linewidth]{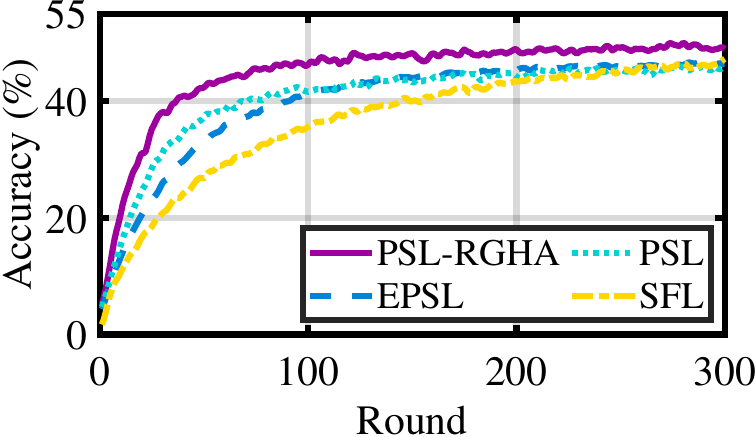}}
    \subfloat[{\fontsize{5.5pt}{6pt}\selectfont ResNet32 FMNIST (non-IID).}]{
        \includegraphics[width=0.16\linewidth]{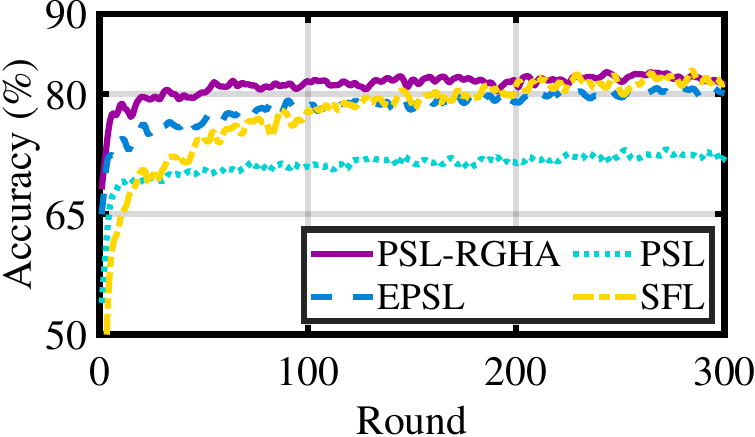}}
    \\[0.5ex]
    \caption{\rmfamily Performance of PSL-RGHA on FMNIST, CIFAR10, and CIFAR100 datasets under IID and non-IID settings.}
    \label{fig:6}
\end{figure*}
}

\begin{figure}[t]
    \captionsetup[subfloat]{skip=-1.5pt}
    \vspace{-7pt} 
    \centering
    \subfloat[\rmfamily \fontsize{7.5pt}{8pt}\selectfont Speech commands (IID).]{
        \includegraphics[width=0.43\linewidth]{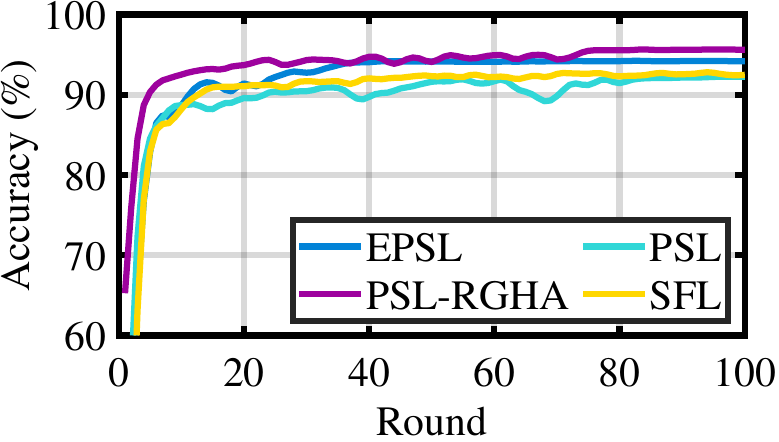}}
        \hspace{0.1\linewidth}
    \subfloat[\rmfamily\fontsize{7.5pt}{8pt}\selectfont SNIPS (IID).]{
        \includegraphics[width=0.43\linewidth]{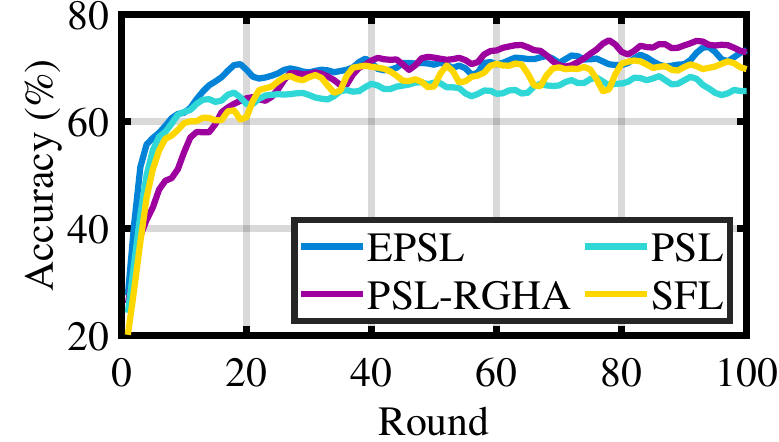}}
    \\[0.5ex]
    \subfloat[\rmfamily \fontsize{7.5pt}{8pt}\selectfont Speech commands (non-IID).]{
        \includegraphics[width=0.43\linewidth]{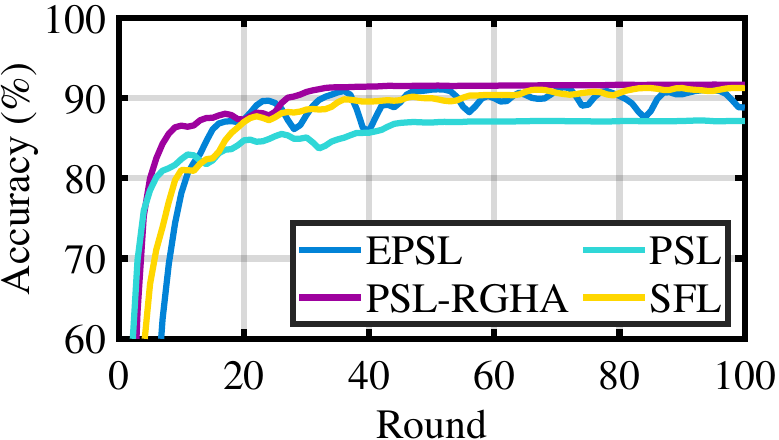}}
        \hspace{0.1\linewidth}
    \subfloat[\rmfamily \fontsize{7.5pt}{8pt}\selectfont SNIPS (non-IID).]{
        \includegraphics[width=0.43\linewidth]{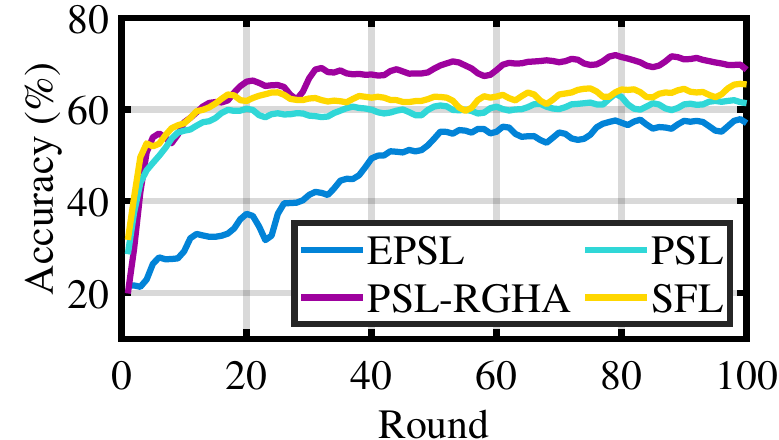}}
    \\[0.5ex]
    \caption{\rmfamily  Performance of PSL-RGHA on Speech commands and SNIPS datasets under IID and non-IID settings.}
    \label{fig:yuyin}
\end{figure}

\begin{proof}
The core of this proof relies on the MM principle. We proceed in three logical steps: constructing a surrogate majorizer function, minimizing this surrogate, and establishing the final descent chain.

We utilize the algebraic inequality $\sqrt{z}\le \frac{z}{2a}+\frac{a}{2}$ (valid for any $z\ge 0$ and $a>0$, with equality if and only if $a=\sqrt{z}$). By substituting $z=\|g-u_{r_\ell}^{t}\|^{2}+\varepsilon^{2}$ and $a=d_\ell\ge \varepsilon>0$, we obtain a localized upper bound for each term in the objective:
\[
\sqrt{\|g-u_{r_\ell}^{t}\|^{2}+\varepsilon^{2}} \le \frac{\|g-u_{r_\ell}^{t}\|^{2}+\varepsilon^{2}}{2d_\ell}+\frac{d_\ell}{2}.
\]
Summing these bounds over all $\ell=1,\ldots,k$ yields the surrogate majorizing function $Q(g \mid g^{(0)})$:
\[
Q(g \mid g^{(0)}) \triangleq \sum_{\ell=1}^{k}\left(\frac{\|g-u_{r_\ell}^{t}\|^{2}+\varepsilon^{2}}{2d_\ell}+\frac{d_\ell}{2}\right).
\]
By construction, this surrogate strictly upper-bounds our objective function $\Phi_{k,\varepsilon}^{t}(g)\le Q(g \mid g^{(0)})$ for all $g\in\mathcal{C}$, and tightens exactly at the current estimate: $Q(g^{(0)} \mid g^{(0)})=\Phi_{k,\varepsilon}^{t}(g^{(0)})$.

Next, we find the optimal update by minimizing $Q(\cdot \mid g^{(0)})$ over the feasible set $\mathcal{C}$. Discarding constants independent of $g$, this is mathematically equivalent to minimizing the weighted least-squares function:
\[
J(g)\triangleq \frac{1}{2}\sum_{\ell=1}^{k} w_\ell \|g-u_{r_\ell}^{t}\|^{2},\qquad g\in\mathcal{C}.
\]
Let $\alpha\triangleq\sum_{\ell=1}^{k} w_\ell > 0$. By completing the square, we can rewrite $J(g)$ as:
\[
J(g)=\frac{\alpha}{2}\|g-\tilde g^{(1)}\|^{2}+\text{const}, \quad \text{where} \quad \tilde g^{(1)}=\frac{\sum_{\ell=1}^{k} w_\ell u_{r_\ell}^{t}}{\sum_{\ell=1}^{k} w_\ell}.
\]
Since each $w_\ell>0$ and $\|u_{r_\ell}^{t}\|=1$, the vector $\tilde g^{(1)}$ is a convex combination of points on the unit sphere. Consequently, $\|\tilde g^{(1)}\|\le 1$, meaning $\tilde g^{(1)}$ already lies perfectly within our constraint set $\mathcal{C}$ (the constraint is inactive). Therefore, the minimizer over $\mathcal{C}$ is exactly $g^{(1)}=\tilde g^{(1)}$. Because $g^{(1)}$ minimizes the majorizer $Q$, we can establish the final descent chain:
\[
\Phi_{k,\varepsilon}^{t}(g^{(1)}) \le Q(g^{(1)} \mid g^{(0)}) \le Q(g^{(0)} \mid g^{(0)}) = \Phi_{k,\varepsilon}^{t}(g^{(0)}).
\]
This confirms the monotonic descent property and completes the proof.
\end{proof}

Establishing Theorem~\ref{thm:mm_weiszfeld} that RGPC reliably constructs a robust global descent direction, we now demonstrate how GHA utilizes this reference to regularize heterogeneous client updates. Theorem~\ref{thm:gha_geom_compact} proves that the GHA projection not only guarantees strict directional alignment with the global trend, but also inherently ensures numerical stability by preventing gradient norm expansion:

\begin{theorem}[Geometric meaning and stability of GHA]
\label{thm:gha_geom_compact}
Let $u\in\mathbb{R}^d$ be a unit vector and define the reference half-space 
$\mathcal{H}\triangleq\{d\in\mathbb{R}^d:\langle d,u\rangle\ge 0\}$. 
For any $g\in\mathbb{R}^d$, let $\hat g$ be given by \eqref{eq:gha_projection}. Then:
\begin{enumerate}[leftmargin=3.1em,itemsep=0.1em,topsep=0.1em]
\item \emph{(Half-space feasibility and monotone reference projection)} 
$\langle \hat g,u\rangle\ge 0$ and $\langle \hat g,u\rangle\ge \langle g,u\rangle$.
\item \emph{(Minimum change under conflict)} 
If $\langle g,u\rangle<0$, then $\hat g=\Pi_{\mathcal{H}}(g)$ and thus 
$\|\hat g-g\|=\min_{d\in\mathcal{H}}\|d-g\|$.
\item \emph{(Non-expansiveness in norm)} 
$\|\hat g\|\le \|g\|$.
\end{enumerate}
\end{theorem}

\begin{proof}
Let $s\triangleq\langle g,u\rangle$ be the projection scalar. We analyze the alignment behavior under two mutually exclusive scenarios based on the gradient conflict. According to the piecewise alignment rule in \eqref{eq:gha_projection}, we have $\hat g=g-su$. Consequently, the inner product with the reference direction becomes:
\[
\langle \hat g,u\rangle = s - s = 0.
\]
This implies both $\langle \hat g,u\rangle\ge 0$ and $\langle \hat g,u\rangle\ge s$, satisfying condition (1). 
Furthermore, since $\hat g\perp u$ and the original update can be decomposed as $g=\hat g + su$, applying the Pythagorean theorem yields:
\[
\|g\|^2=\|\hat g\|^2+s^2 \quad \Rightarrow \quad \|\hat g\|\le \|g\|.
\]
Regarding the projection claim (2), the Euclidean projection onto the half-space $\mathcal{H}$ satisfies $\Pi_{\mathcal{H}}(g)=g$ if $s\ge 0$, and $\Pi_{\mathcal{H}}(g)=g-su$ if $s<0$. Thus, $\hat g=\Pi_{\mathcal{H}}(g)$, and the minimum-distance property naturally follows from the standard definition of Euclidean projections.

Let $\lambda \in [0, 1/2]$ be the alignment coefficient defined in \eqref{eq:gha_projection}, such that the softly aligned gradient is $\hat g=(1-\lambda)g+\lambda\|g\|u$. The inner product with the reference direction expands as:
\[
\begin{aligned}
\langle \hat g,u\rangle 
&=(1-\lambda)s+\lambda\|g\| \\
&\ge (1-\lambda)s+\lambda s \;=\; s \;\ge\; 0,
\end{aligned}
\]
which proves condition (1) for this case.

To establish the non-expansiveness property (3), we introduce auxiliary variables $a\triangleq g$ and $b\triangleq \|g\|u$. Notice that $\|a\|=\|b\|=\|g\|$, and our aligned gradient can be written as the convex combination $\hat g=(1-\lambda)a+\lambda b$. Using the standard norm identity for affine combinations:
\[
\|(1-\lambda)a+\lambda b\|^2 = (1-\lambda)\|a\|^2+\lambda\|b\|^2-\lambda(1-\lambda)\|a-b\|^2,
\]
we can substitute $a$ and $b$ back to obtain:
\[
\|\hat g\|^2 = \|g\|^2-\lambda(1-\lambda)\|g-\|g\|\,u\|^2 \le \|g\|^2.
\]
Since $\lambda(1-\lambda) \ge 0$ for $\lambda \in [0, 1/2]$, this immediately yields $\|\hat g\|\le \|g\|$. Combining both cases effectively covers the entire gradient space, completing the proof.
\end{proof}


\section{Experimental}

\subsection{Experimental Setup}
To comprehensively evaluate the performance of PSL-RGHA, we utilize five standard benchmarks: FMNIST~\cite{19}, CIFAR-10, CIFAR-100~\cite{20}, Speech commands~\cite{Speech} and SNIPS~\cite{snips} under both IID and non-IID settings. 
We employ ResNet-18 and ResNet-32~\cite{21} as the backbone architectures. 
To align with the split learning paradigm, each model is partitioned at a specific cut layer, with the initial layers deployed on the client and the remaining layers on the server. The training environment consists of $10$ client devices.
To simulate statistical heterogeneity (non-IID data), we partition the training data among clients using a Dirichlet distribution $Dir(\alpha)$~\cite{22}, where the concentration parameter $\alpha$ is varied across $\{0.1, 0.3, 0.5, 0.7\}$ to represent increasing degrees of data skew. 
All models are trained by the SGD~\cite{22-1} optimizer.

\begin{table}[t]
\vspace{-7pt} 
\centering
\small
\caption{Test accuracy (\%) of PSL-RGHA and baselines under IID and non-IID settings.}
\vspace{-1.5pt}
\setlength{\tabcolsep}{4pt}
\renewcommand{\arraystretch}{0.9}
\resizebox{\linewidth}{!}{
\begin{tabular}{c c c c c c c}
\toprule
Dataset & Setting & Backbone &  PSL & SFL & EPSL & \textbf{PSL-RGHA} \\
\midrule

\multirow{4}{*}{CIFAR10}
& \multirow{2}{*}{IID} 
&  ResNet-18 &  78.36    & 83.71 & 84.62 & \textbf{84.98}  \\
&  &  ResNet-32  & 83.71 & 86.11 & 86.41 & \textbf{87.66}  \\
\cmidrule{2-7}
& \multirow{2}{*}{non-IID} 
&  ResNet-18 &  52.32    & 79.74 & 67.01 & \textbf{82.68}  \\
&  &  ResNet-32  & 70.26 & 75.08 & 75.61 & \textbf{79.17}  \\

\midrule

\multirow{4}{*}{CIFAR100}
& \multirow{2}{*}{IID}
&  ResNet-18 &  56.42    & 46.52 & 54.08 & \textbf{59.28}  \\
&  &  ResNet-32  & 58.04 & 59.88 & 58.62 & \textbf{60.23}  \\
\cmidrule{2-7}
& \multirow{2}{*}{non-IID} 
&  ResNet-18 &  44.87    & 39.71 & 45.71 & \textbf{47.52}  \\
&  &  ResNet-32  & 46.14 & 47.61 & 46.84 & \textbf{49.58}  \\
\midrule

\multirow{4}{*}{FMNIST}
& \multirow{2}{*}{IID}
&  ResNet-18 &  89.17    & 90.95 & 89.21 & \textbf{93.49}  \\
&  &  ResNet-32  & 87.72 & 89.52 & 90.41 & \textbf{91.25}  \\
\cmidrule{2-7}
& \multirow{2}{*}{non-IID}
&  ResNet-18 &  78.47   & 80.42 & 81.36 & \textbf{86.83}  \\
&  &  ResNet-32  & 72.89 & 82.37 & 81.55 & \textbf{83.05}  \\

\bottomrule
\end{tabular}
}
\vspace{0em}

\label{tab:1}
\vspace{0em}
\end{table}

\subsection{Benchmarks}
To comprehensively evaluate the performance of PSL-RGHA, we compare the proposed PSL-RGHA against the following representative baselines: 
i) SFL~\cite{6} is a hybrid architecture that integrates SL with FL. 
SFL enables parallel client-side training across client devices, where each device updates its client-side model and uploads it to the parameter server for model aggregation. ii) PSL~\cite{8} is a streamlined variant of SFL that eliminates client-side model aggregation. PSL trains a shared server-side model using activations from all clients, while each client independently updates its local sub-model without cross-device coordination. iii) EPSL~\cite{23} is optimized PSL variant that reduces computational and communication burdens by aggregating only a fraction of the cut-layer gradients on the server side, providing a middle ground between full aggregation and total independence.

\subsection{Performance Evaluation}

\subsubsection{Superiority of PSL-RGHA}

We first evaluate the convergence and generalization performance of PSL-RGHA. Fig.~\ref{fig:6} and Fig.~\ref{fig:yuyin} show the test accuracy curves on vision and speech tasks, respectively, while Table~\ref{tab:1} reports the final top-1 accuracy under different datasets and heterogeneity settings. PSL-RGHA consistently achieves faster convergence and higher accuracy than all baselines across vision, keyword spotting, and speech classification tasks. This gain mainly comes from its gradient guidance and correction mechanism. Unlike vanilla PSL, which suffers from severe client drift, PSL-RGHA uses GHA to project conflicting gradients onto the reference half-space, ensuring constructive updates and reducing oscillations caused by data heterogeneity. EPSL only performs partial aggregation and therefore cannot fully correct gradient conflicts, resulting in lower convergence performance.

\begin{figure}[t]
    \captionsetup[subfloat]{skip=-2pt}
    \vspace{-3pt} 
    \centering
    \subfloat[\rmfamily\tiny CIFAR-10 (IID).]{
        \includegraphics[width=0.32\linewidth]{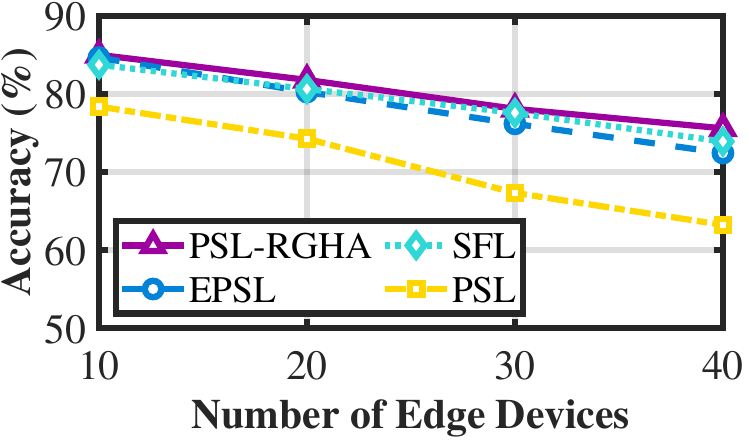}}
    \subfloat[\rmfamily\tiny CIFAR-100 (IID).]{
        \includegraphics[width=0.32\linewidth]{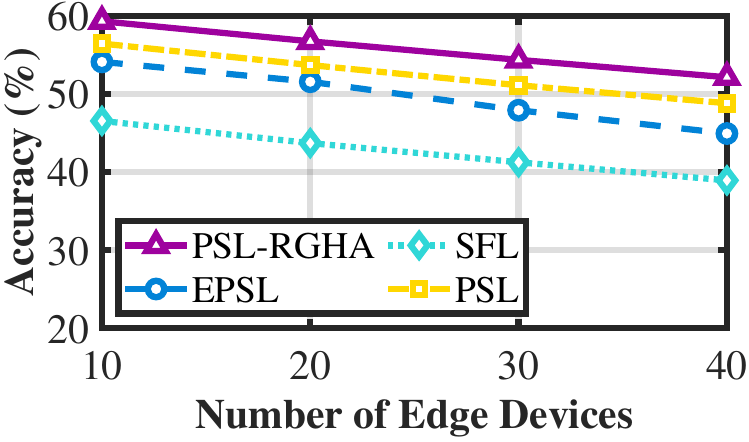}}
    \subfloat[\rmfamily\tiny FMNIST (IID).]{
        \includegraphics[width=0.32\linewidth]{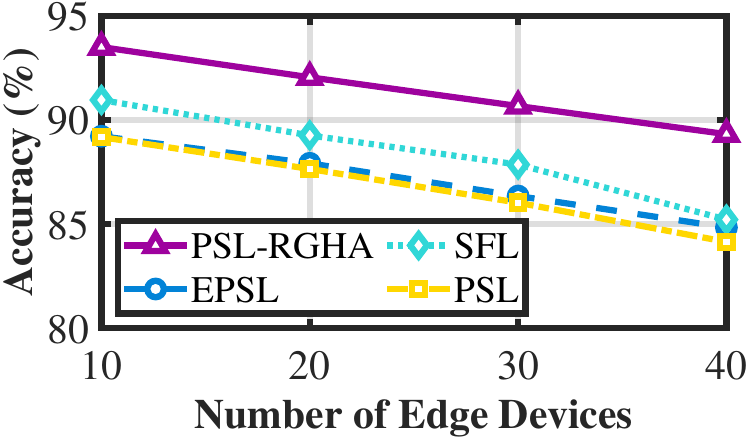}}
    \\[0.5ex]
    \vspace{-3pt}
    \subfloat[\rmfamily\tiny CIFAR-10 (non-IID).]{
        \includegraphics[width=0.32\linewidth]{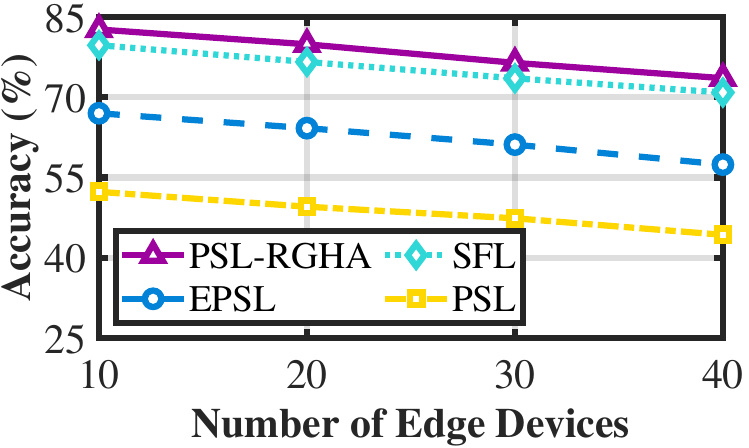}}
    \subfloat[\rmfamily\tiny CIFAR-100 (non-IID).]{
        \includegraphics[width=0.32\linewidth]{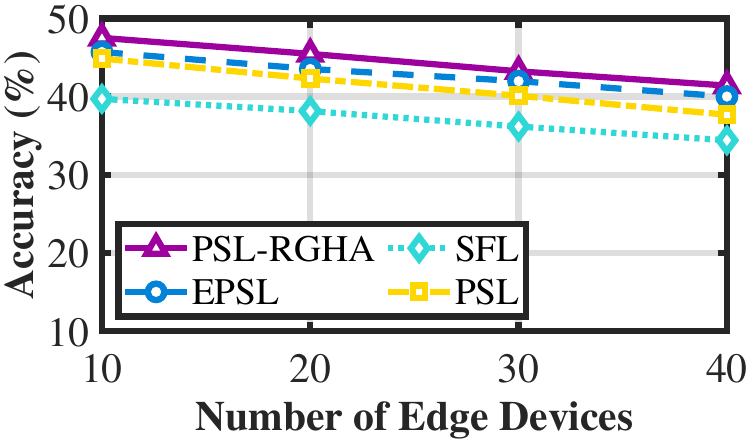}}
    \subfloat[\rmfamily\tiny FMNIST (non-IID).]{
        \includegraphics[width=0.32\linewidth]{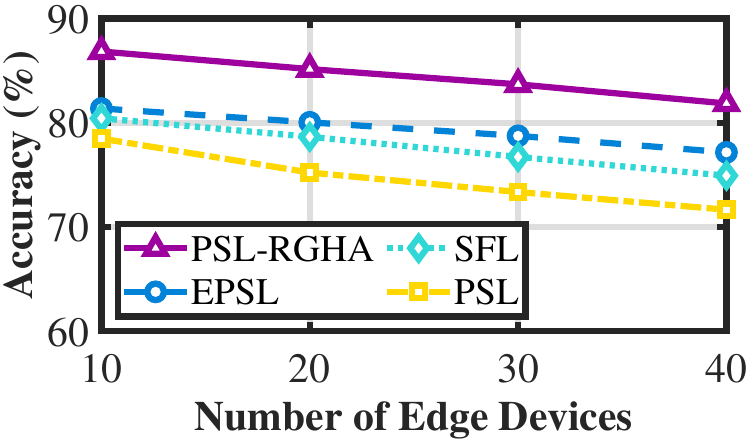}}
    \\[0.5ex]
    \vspace{0pt}
    \caption{\rmfamily The impact of the number of edge devices on ResNet-18 test accuracy under IID and non-IID settings.}
    \label{fig:7}
\end{figure}

\begin{figure}[t]
    \captionsetup[subfloat]{skip=-2pt}
    \vspace{-3pt} 
    \centering
    \subfloat[\rmfamily\tiny ResNet-18 CIFAR-10.]{
        \includegraphics[width=0.32\linewidth]{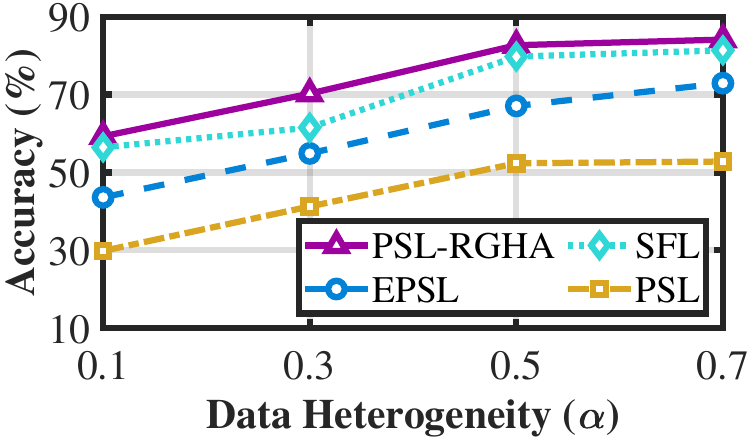}}
    \subfloat[\rmfamily\tiny ResNet-18 CIFAR-100.]{
        \includegraphics[width=0.32\linewidth]{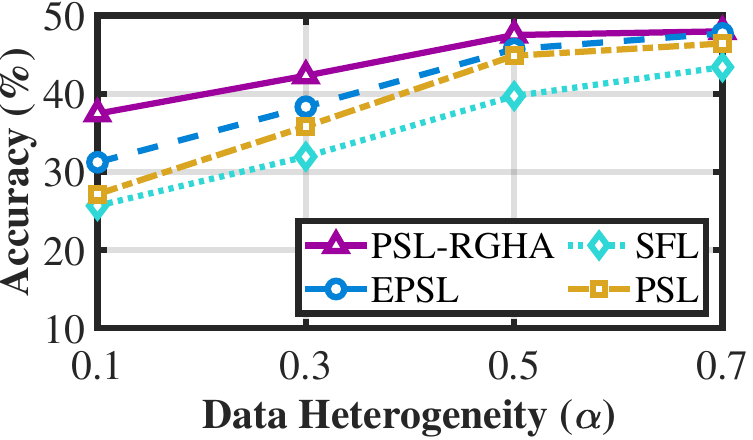}}
    \subfloat[\rmfamily\tiny ResNet-18 FMNIST.]{
        \includegraphics[width=0.32\linewidth]{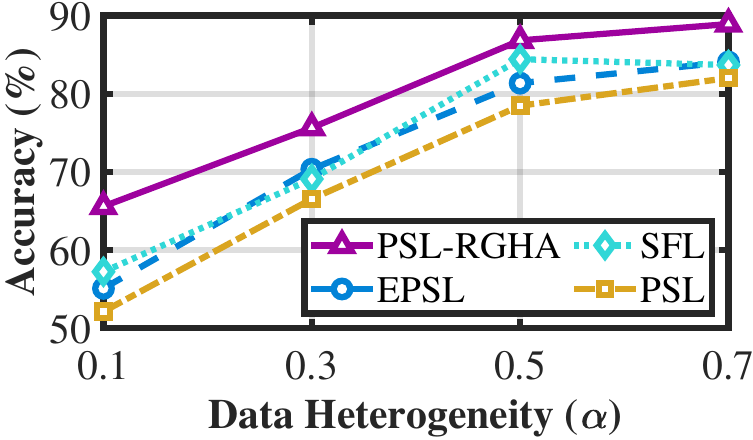}}
    \\[0.5ex]
      \vspace{-3pt}
    \subfloat[\rmfamily\tiny ResNet-32 CIFAR-10.]{
        \includegraphics[width=0.32\linewidth]{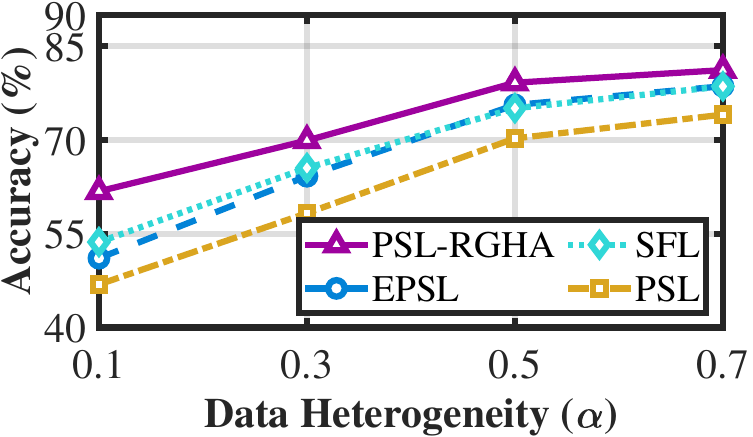}}
    \subfloat[\rmfamily\tiny ResNet-32 CIFAR-100.]{
        \includegraphics[width=0.32\linewidth]{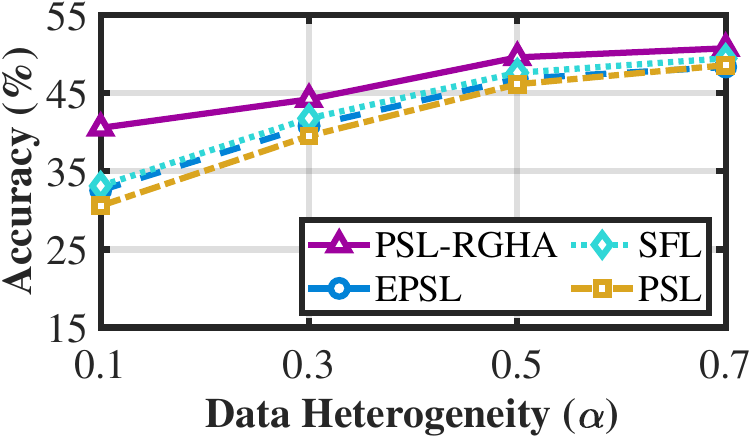}}
    \subfloat[\rmfamily\tiny ResNet-32 FMNIST.]{
        \includegraphics[width=0.32\linewidth]{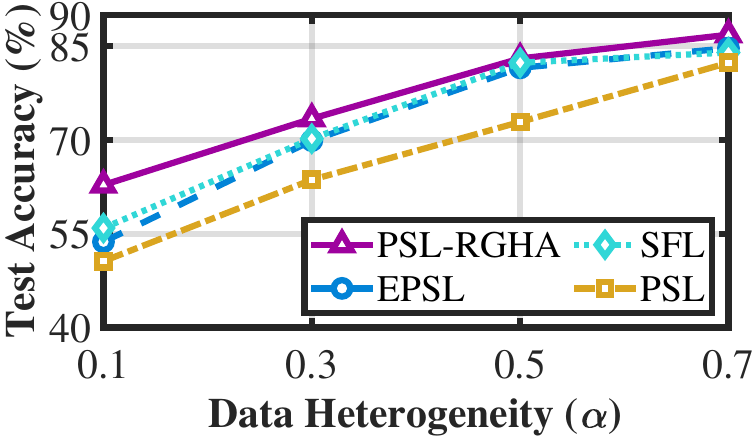}}
    \\[0.5ex]
    \vspace{0pt}
    \caption{\rmfamily The impact of data heterogeneity on test accuracy under non-IID settings.}
    \vspace{-5pt}
    \label{fig:8}
\end{figure}

\subsubsection{Impact of Client Scalability}
Fig.~\ref{fig:7} illustrates the converged accuracy as the number of participating clients varies from $10$ to $40$. 
PSL-RGHA shows superior scalability, consistently outperforming baselines across all datasets and settings.
In balanced IID environments, all methods remain relatively stable as the system scales. 
However, PSL-RGHA maintains a distinct performance edge. 
This stems from the GHA module, which refines the descent trajectory even when data distributions are uniform, while the other baselines are limited by the inherent variance of averaging stochastic gradients. 
Under heterogeneous settings, as the client scale increases, baselines suffer from significant performance deterioration. 
This is attributed to the accumulation of gradient noise.
With naive aggregation, adding more heterogeneous clients introduces more conflicting update directions, which dilute the global signal. 
Conversely, PSL-RGHA demonstrates remarkable robustness, confirming that the RGPC module effectively filters out the tail noise introduced by additional clients, ensuring that the reference direction remains robust regardless of the system scale.

\subsubsection{Impact of Data Heterogeneity}
Fig.~\ref{fig:8} evaluates the model robustness against statistical heterogeneity.
As the heterogeneity increase ($\alpha$ decreases from $0.7$ to $0.1$), all baseline methods suffer from performance degradation. 
This confirms that standard aggregation mechanisms are highly sensitive to the gradient bias induced by skewed data distributions.
In contrast, PSL-RGHA exhibits exceptional statistical resilience. 
Even under extreme non-IID settings ($\alpha=0.1$), it maintains superior accuracy with only marginal degradation compared to the IID scenarios. 
This robustness validates that the proposed GHA mechanism effectively decouples local data bias from the global optimization direction, enabling the model to adapt to severe distribution shifts.

\subsubsection{Ablation Study}
\begin{figure}[t]
    \captionsetup[subfloat]{skip=-2pt}
    \vspace{-3pt} 
    \centering
    \subfloat[\rmfamily\tiny  CIFAR-10(IID).]{
        \includegraphics[width=0.32\linewidth]{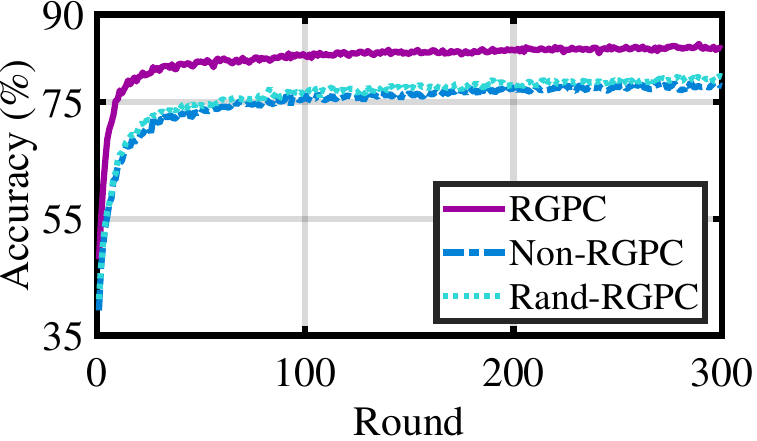}}
    \subfloat[\rmfamily\tiny  CIFAR-100(IID).]{
        \includegraphics[width=0.32\linewidth]{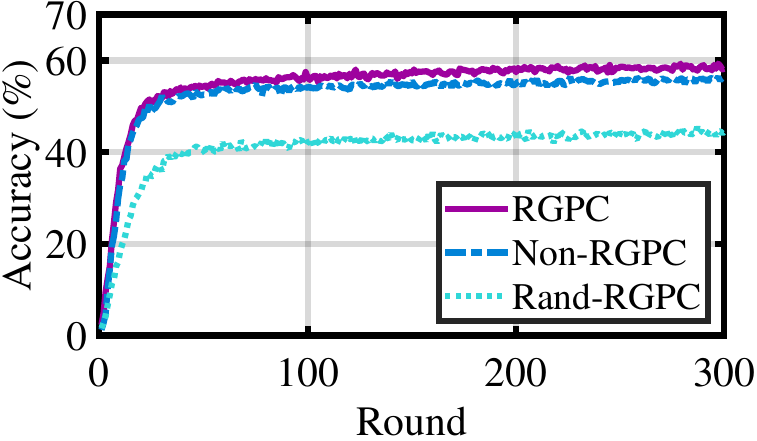}}
    \subfloat[\rmfamily\tiny  FMNIST(IID).]{
        \includegraphics[width=0.32\linewidth]{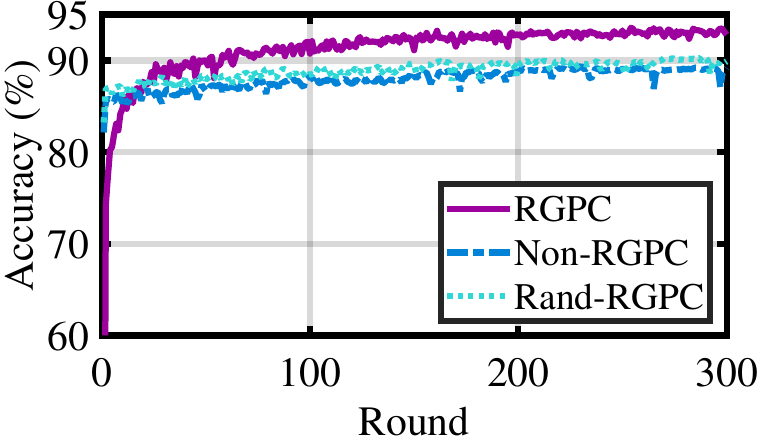}}
    \\[1ex]
    \vspace{-3pt}
    \subfloat[\rmfamily\tiny  CIFAR-10(non-IID).]{
        \includegraphics[width=0.32\linewidth]{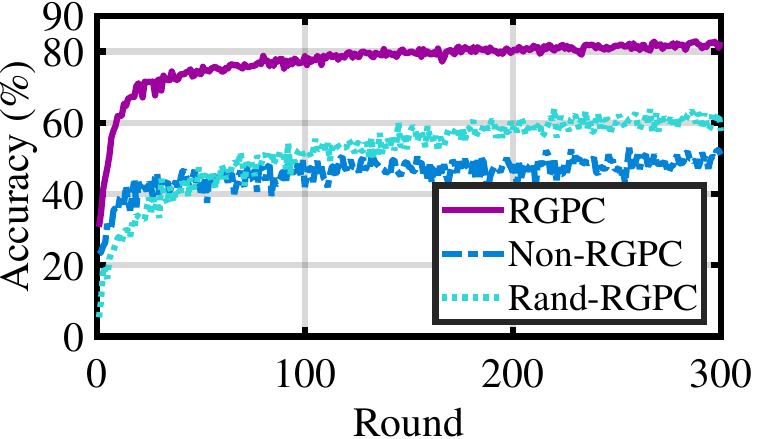}}
    \subfloat[\rmfamily\tiny  CIFAR-100(non-IID).]{
        \includegraphics[width=0.32\linewidth]{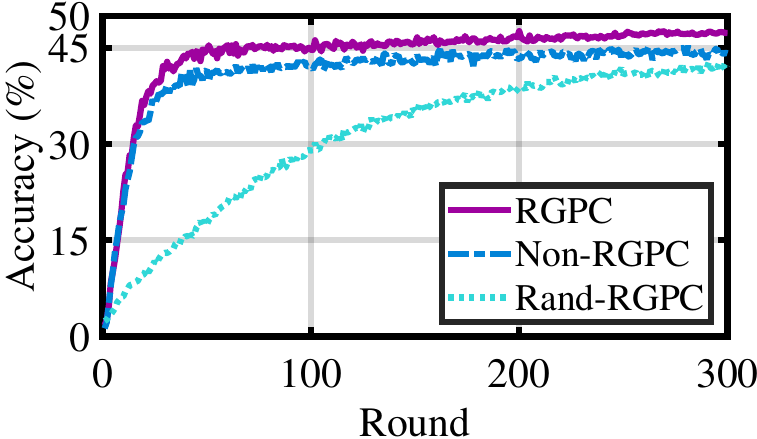}}
    \subfloat[\rmfamily\tiny  FMNIST(non-IID).]{
        \includegraphics[width=0.32\linewidth]{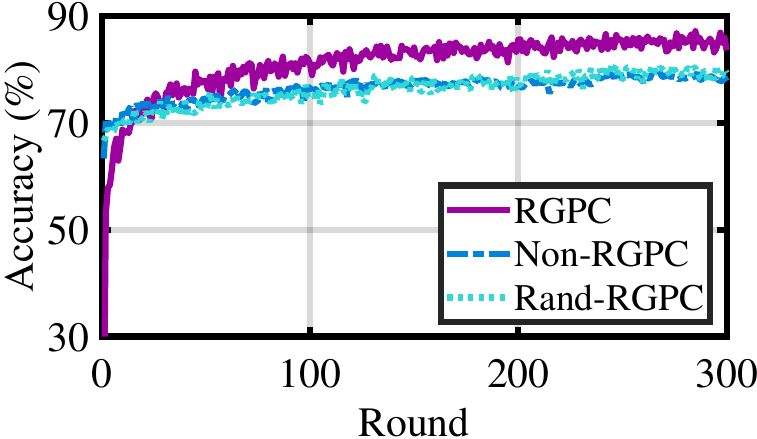}}
    \\[0.5ex]
    \vspace{1pt}
    \caption{\rmfamily Ablation study of RGPC on CIFAR-10, CIFAR-100, and FMNIST under IID and non-IID settings (ResNet-18).}
    \label{fig:9}
\end{figure}

We first isolate the contribution of the RGPC module by comparing it against two variants: (i) Non-RGPC (Naive Strategy), which omits the robust construction steps, and (ii) Rand-RGPC (Stochastic Strategy), which constructs the reference from a random subset of client gradients. 
As shown in Fig.~\ref{fig:9}, RGPC consistently outperforms both variants. The superiority over Rand-RGPC reveals that reference quality is critical: relying on random subsets introduces stochastic noise that destabilizes the alignment anchor. Similarly, the gap over Non-RGPC confirms that without the Weiszfeld-based filtering, the reference direction is easily skewed by outliers in non-IID data. 
RGPC effectively provides a stable anchor for the subsequent alignment.

\begin{figure}[t]
    \captionsetup[subfloat]{skip=-2pt}
    \vspace{-3pt} 
    \centering
    \subfloat[\rmfamily\tiny CIFAR-10 (IID).]{
        \includegraphics[width=0.32\linewidth]{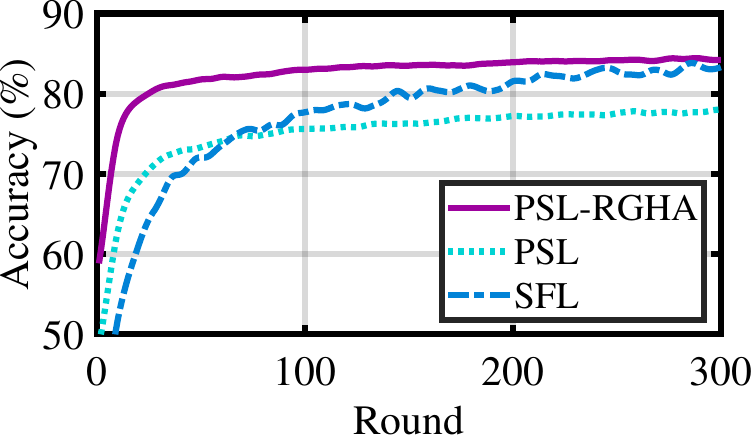}}
    \subfloat[\rmfamily\tiny CIFAR-100 (IID).]{
        \includegraphics[width=0.32\linewidth]{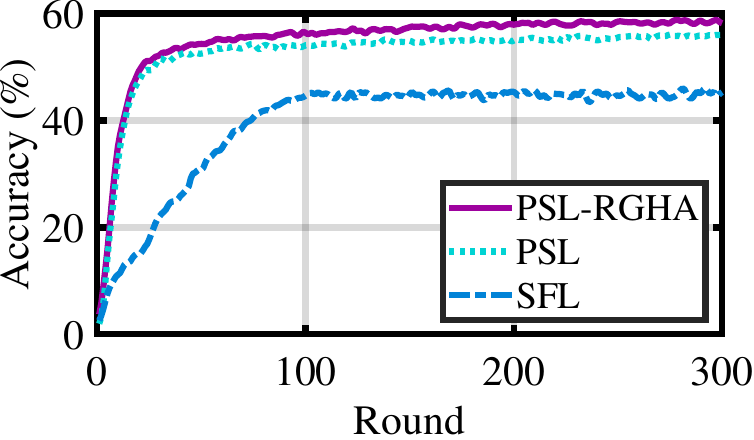}}
    \subfloat[\rmfamily\tiny FMNIST (IID).]{
        \includegraphics[width=0.32\linewidth]{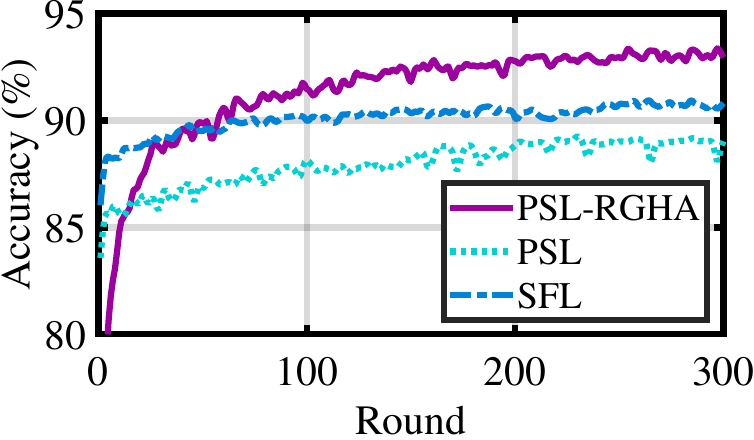}}
    \\[0.5ex]
    \vspace{-3pt}
    \subfloat[\rmfamily\tiny  CIFAR-10(non-IID).]{
        \includegraphics[width=0.32\linewidth]{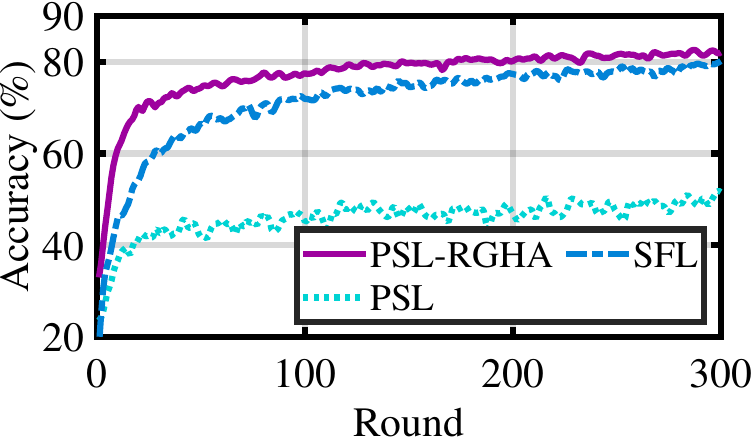}}
    \subfloat[\rmfamily\tiny CIFAR-100(non-IID).]{
        \includegraphics[width=0.32\linewidth]{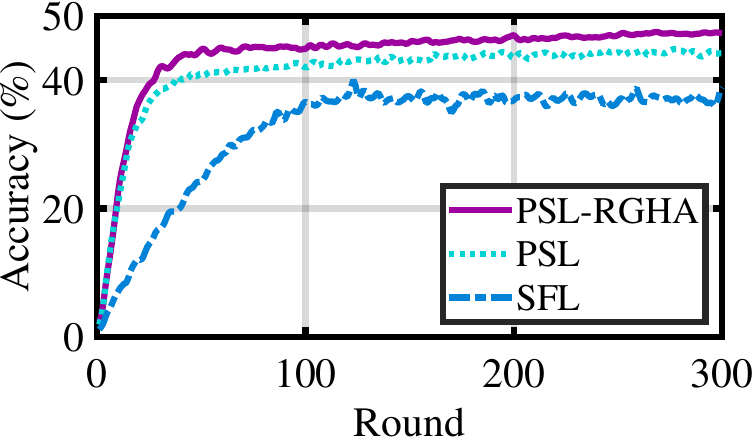}}
    \subfloat[\rmfamily\tiny FMNIST(non-IID).]{
        \includegraphics[width=0.32\linewidth]{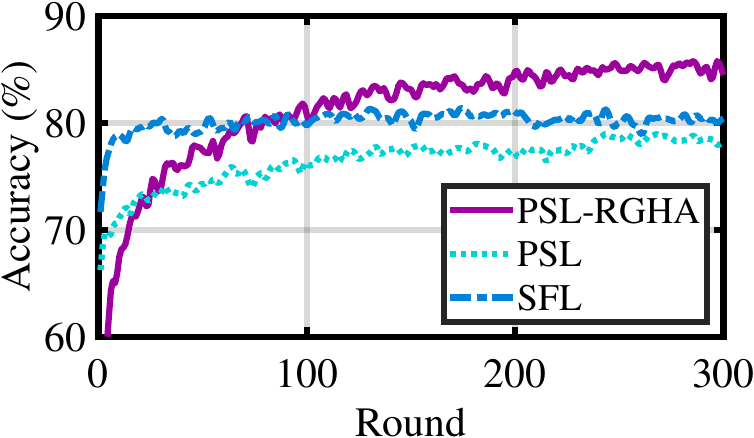}}
    \\[0.5ex]
    \vspace{0pt}
    \caption{\rmfamily Ablation study of GHA on CIFAR-10, CIFAR-100, and FMNIST under IID and non-IID settings (ResNet-18).}
    \vspace{-5pt}
    \label{fig:10}
\end{figure}

We then verify the necessity of the GHA mechanism by examining the performance gap among full PSL-RGHA framework, baseline PSL (without alignment) and SFL (naive aggregation), as illustrated in Fig.~\ref{fig:10}. 
Experimental results demonstrate that GHA yields substantial accuracy gains, confirming that simply aggregating gradients (SFL) or allowing independent updates (PSL) is insufficient under heterogeneity. 
The effectiveness of GHA stems from its ability to actively rectify gradient conflicts. 
By projecting conflicting updates onto the reference half-space, GHA eliminates destructive interference while preserving constructive local features, thereby converting optimization drift into stable convergence.

\section{Conclusion}
In this paper, we proposed PSL-RGHA to resolve the fundamental conflict between client-side parallelism and training stability in SL. Unlike traditional approaches that rely on naive model averaging, PSL-RGHA introduces a geometric perspective to distributed optimization. Through RGPC and GHA, the framework actively filters out directional noise and rectifies divergent updates, decoupling local data bias from the global optimization trajectory. Extensive experiments validate that PSL-RGHA achieves superior performance, proving that explicit gradient alignment is crucial for robust learning on non-IID data. These findings open new avenues for designing aggregation-free distributed learning systems. Moving forward, we aim to extend this geometric alignment principle to asynchronous training scenarios and larger-scale foundation model fine-tuning.

\begin{acks}
This work was supported by the Fujian Provincial Science and Technology Plan Project under Grant No. 2024H6020, and the Green and Intelligent Ship Project in Fujian Province under Grant No. CBG4N21-4-4.
\end{acks}

\bibliographystyle{ACM-Reference-Format}
\bibliography{ACMMM26}

\end{document}